\documentclass[sigconf]{acmart}

\AtBeginDocument{%
  \providecommand\BibTeX{{%
    \normalfont B\kern-0.5em{\scshape i\kern-0.25em b}\kern-0.8em\TeX}}}

\usepackage{multirow}
\usepackage{color}
\usepackage{graphics}
\usepackage{bbding}
\def\etal{\emph{et al. }}

\renewcommand\footnotetextcopyrightpermission[1]{} % removes footnote with conference information in first column
\begin{document}
\fancyhead{}

\title{Question-controlled Text-aware Image Captioning}

\author{Anwen Hu}
\affiliation{%
  \institution{School of Information\\ Renmin University of China}
  \country{}
  }
\email{anwenhu@ruc.edu.cn}

\author{Shizhe Chen}
% \authornote{This work was performed when Shizhe Chen was at Renmin University of China.}
\affiliation{%
 \institution{INRIA}
 \country{}
}
\email{shizhe.chen@inria.fr}

\author{Qin Jin}
\authornote{Corresponding Author}
\affiliation{%
  \institution{School of Information\\ Renmin University of China}
  \country{}
  }
\email{qjin@ruc.edu.cn}

\renewcommand{\shortauthors}{Hu et al.}

\begin{abstract}
For an image with multiple scene texts, different people may be interested in different text information. Current text-aware image captioning models are not able to generate distinctive captions according to various information needs. To explore how to generate personalized text-aware captions, we define a new challenging task, namely Question-controlled Text-aware Image Captioning (Qc-TextCap). With questions as control signals, this task requires models to understand questions, find related scene texts and describe them together with objects fluently in human language. Based on two existing text-aware captioning datasets, we automatically construct two datasets, ControlTextCaps and ControlVizWiz to support the task. We propose a novel Geometry and Question Aware Model (GQAM). GQAM first applies a Geometry-informed Visual Encoder to fuse region-level object features and region-level scene text features with considering spatial relationships. Then, we design a Question-guided Encoder to select the most relevant visual features for each question. Finally, GQAM generates a personalized text-aware caption with a Multimodal Decoder. Our model achieves better captioning performance and question answering ability than carefully designed baselines on both two datasets. With questions as control signals, our model generates more informative and diverse captions than the state-of-the-art text-aware captioning model. Our code and datasets are publicly available at \url{https://github.com/HAWLYQ/Qc-TextCap}. 
\end{abstract}





\maketitle

\section{Introduction}
\begin{figure}
    \centering
    \includegraphics[width=0.9\linewidth]{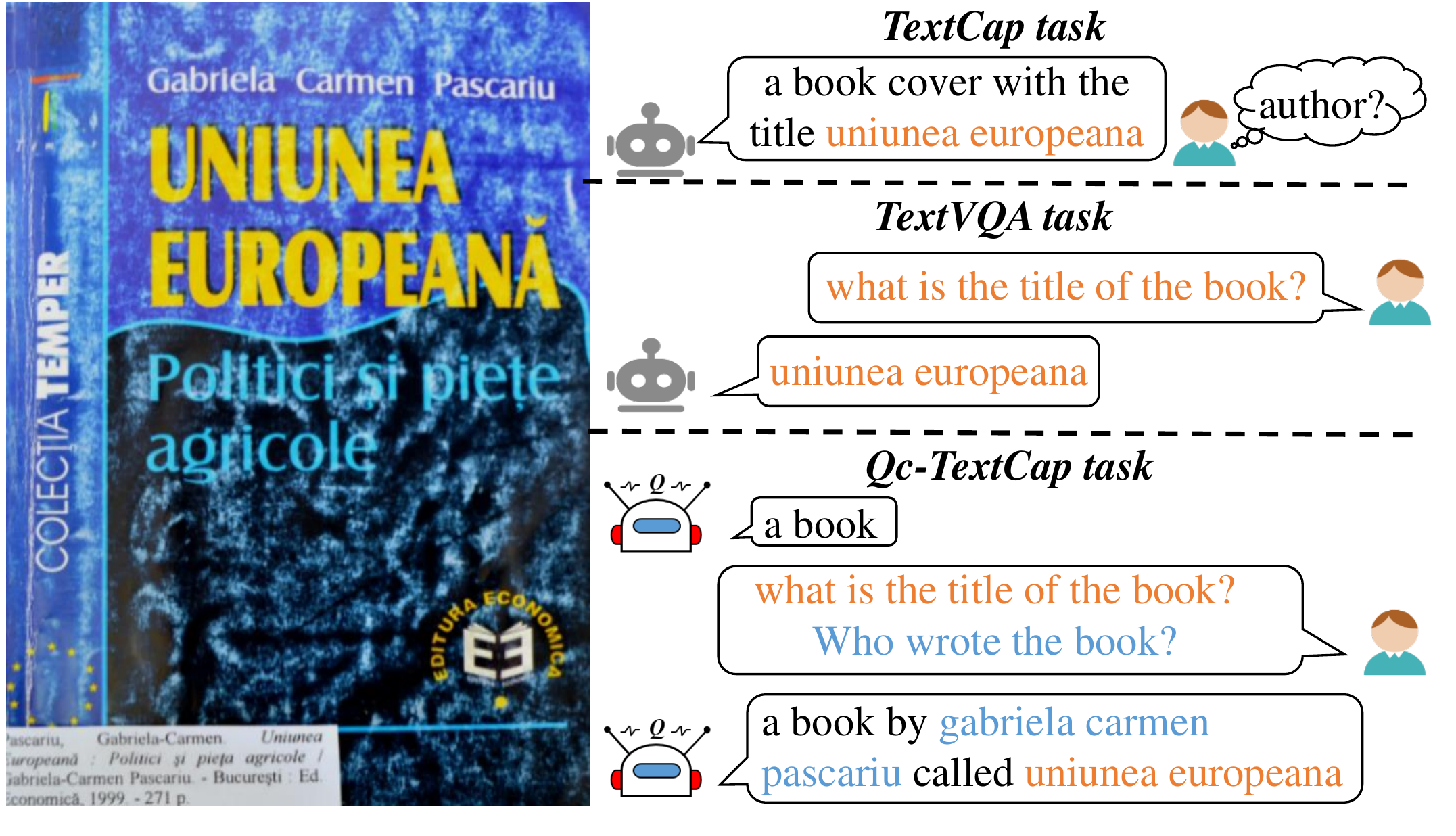}
    \caption{An example of Question-controlled Text-aware Image Captioning (Qc-TextCap). The corresponding question and answer are labelled with the same color.}
    \label{fig:example}
\end{figure}

Texts are omnipresent and convey valuable information of the visual environment, such as the title of a book, the time shown on a clock and the words on a road sign. 
With the goal of describing the visual world to visually impaired people, it is essential to comprehend such scene texts beyond pure visual recognition \cite{rennie2017self,DBLP:conf/icml/XuBKCCSZB15,DBLP:conf/cvpr/00010BT0GZ18,DBLP:conf/cvpr/LuXPS17,DBLP:conf/cvpr/VinyalsTBE15,DBLP:conf/iccv/HuangWCW19,DBLP:conf/emnlp/FischLCCB20}.
Therefore, more recent works are focusing on the text-aware image captioning task \cite{DBLP:conf/aaai/WangBWL21,DBLP:conf/aaai/ZhuGWW21,DBLP:conf/mm/WangTL20,yang2020tap,DBLP:conf/eccv/SidorovHRS20}, which aims to describe an image in natural sentences covering scene text information in the image.

However, when an image consists of rich scene text information as shown in Figure~\ref{fig:example}, it can be tedious and often not necessary to describe all of the texts in the image.
According to a user study in \cite{DBLP:conf/chi/MorrisJBC18}, visually impaired people prefer to know their surrounding environment i.e. images in a progressive manner.
For example in Figure~\ref{fig:example}, an image captioning system is more preferable to firstly tell the visually impaired users an overview description of the image i.e. ``a book'', and then let the user interact with the system to obtain more specific details about their interested scene texts, such as ``who wrote the book'' or ``what is the title of the book''.
In this way, the visually impaired users can obtain more personalized text-aware captions according to their interests.

To satisfy such user needs, in this work, we propose a novel task called Question-controlled Text-aware Image Captioning (Qc-TextCap), where users can ask scene text related questions based on an initial coarse-grained image description to obtain a more informative text-aware image description for the questions. 
Compared to the Text Visual Question Answering (TextVQA) task~\cite{DBLP:conf/cvpr/SinghNSJCBPR19,DBLP:conf/iccv/BitenTMBRJVK19,DBLP:conf/cvpr/HuSDR20,DBLP:conf/eccv/KantBASPLA20} which only contains a single question and a single word answer without initial caption, our Qc-TextCap task demands higher multimodal comprehension abilities. For example, the user can simultaneously ask multiple questions and the model should organize all scene text answers associated with their corresponding visual objects in a fluent sentence description as shown in Figure~\ref{fig:example}.
Compared to previous endeavors in controllable image captioning such as providing object tokens \cite{DBLP:conf/cvpr/ZhengLW19}, image regions \cite{DBLP:conf/cvpr/CorniaBC19}, and abstract scene graphs \cite{DBLP:conf/cvpr/ChenJWW20} to generate different captions, using questions as the control signal is a more convenient and natural interaction interface for the visually impaired people, Whereas previous control signals require users to specify which visual objects or regions he/she wants to know.

As existing image caption datasets and VQA datasets do not directly support the Qc-TextCap task, we propose an automatic approach to construct data in the form of quadruples, <image, initial simple caption, control questions, text-aware captions>, from existing text-aware caption datasets i.e. TextCaps \cite{DBLP:conf/eccv/SidorovHRS20} and VizWiz-Captions \cite{DBLP:conf/eccv/GurariZZB20}.
Specifically, given annotated text-aware captions for images in existing datasets, we use automatic approaches to generate an initial caption and questions related to scene texts in the image.
Based on the Qc-TextCap dataset, we further propose a Geometry and Question Aware Model (GQAM) for the Question-controlled Image Captioning task. The GQAM consists of three key modules, namely a Geometry-informed Visual Encoder that fuses object region features with scene text region features considering their spatial relationship; a Question-guided Encoder that attends to corresponding relevant visual features to encode word-level question information; and a Multimodal Decoder that takes the visual region features, question features and initial caption features as input to generate a text-aware caption.
Experimental results on two datasets demonstrate that our model is able to effectively generate text-aware captions to answer different questions.

The main contributions of our work are as follows:
\begin{itemize}
    \item We propose a novel challenging task, namely Question-controlled Text-aware Image Captioning (Qc-TextCap), towards generating informative and personalized image captions to benefit visually-impaired users.
    \item We develop an automatic system to construct two appropriate datasets for the Qc-TextCap task based on existing text-aware image captioning datasets.
    \item We propose a novel captioning model GQAM that progressively encodes relevant multimodal features with Geometry-informed Visual Encoder and Question-guided Encoder and generates informative captions via a Multimodal Decoder.
    \item GQAM outperforms carefully designed baselines in Qc-TextCap and generates more informative and diverse captions than the text-aware captioning model.   
\end{itemize}

\section{Related Work}
\noindent\textbf{General Image Captioning.}
Lots of neural network based models\cite{rennie2017self,DBLP:conf/icml/XuBKCCSZB15,DBLP:conf/cvpr/00010BT0GZ18,DBLP:conf/cvpr/LuXPS17,DBLP:conf/cvpr/VinyalsTBE15,DBLP:conf/iccv/HuangWCW19,DBLP:conf/emnlp/FischLCCB20,DBLP:conf/nips/HerdadeKBS19} have been proposed for general image captioning. AoANet \cite{DBLP:conf/iccv/HuangWCW19} achieves state-of-the-art performance with an attention on attention mechanism.
Fishch \etal \cite{DBLP:conf/emnlp/FischLCCB20} propose to use question answering accuracy as reward during training to increase the amount of information in generated captions. During inference, their method still simply generates a single caption without questions as control signals. 
These methods are able to enumerate major objects and describe their relationships but fail to provide detailed text information in the image.

\noindent\textbf{Text-aware Image Captioning.}
Text-aware Image Captioning aims to comprehend text information in an image and relate it to visual objects. TextCaps \cite{DBLP:conf/eccv/SidorovHRS20} and VizWiz-Captions \cite{DBLP:conf/eccv/GurariZZB20} are two available datasets. Images of TextCaps come from Open Images V3 dataset and are all verified to contain scene texts. Images in VizWiz-Captions are taken by the blind and around 63\% images include scene texts. Images taken by the blind may have quality issues, such as overexposure, but can better represent the real use case for visually-impaired people.
Based on these two datasets, there have been some models \cite{DBLP:conf/aaai/WangBWL21,DBLP:conf/aaai/ZhuGWW21,DBLP:conf/mm/WangTL20,yang2020tap} proposed to improve the quality of text-aware captions. Wang \etal \cite{DBLP:conf/mm/WangTL20} propose to encode intrinsic spatial relationship between OCR tokens to generate more complete scene text information. Zhu \etal \cite{DBLP:conf/aaai/ZhuGWW21} propose a simple strong baseline which consists of multiple attention blocks. Wang \etal \cite{DBLP:conf/aaai/WangBWL21} introduce confidence embedding of OCR tokens to help select the most noteworthy scene texts. Yang \etal \cite{yang2020tap} design text-aware pre-training tasks to enhance the model ability in reading and understanding scene texts.
These works contribute to generating a better text-aware caption for each image but have not explored how to generate personalized captions.

\noindent\textbf{Text Visual Question Answering.}
Models for TextVQA\cite{DBLP:conf/cvpr/SinghNSJCBPR19,DBLP:conf/iccv/BitenTMBRJVK19,DBLP:conf/cvpr/HuSDR20,DBLP:conf/eccv/KantBASPLA20} are designed to find correct scene texts from images to answer questions. There are two major differences between TextVQA task and Qc-TextCap task. First, Qc-TextCap requires models to process multiple questions simultaneously. Second, models for Qc-TextCap should be able to organize multiple scene text answers and relevant objects for a fluent description.

\noindent\textbf{Controllable Image Captioning.}
Controllable Image Captioning task aims to generate captions in different styles or capturing different contents in the image \cite{DBLP:conf/cvpr/ZhengLW19, DBLP:conf/cvpr/CorniaBC19,DBLP:conf/cvpr/ChenJWW20}. Zheng \etal \cite{DBLP:conf/cvpr/ZhengLW19} propose to describe an image with a guiding object as the initial token. Cornia \etal \cite{DBLP:conf/cvpr/CorniaBC19} use image regions as control signals to produce multiple captions for a given image. Chen \etal \cite{DBLP:conf/cvpr/ChenJWW20} further design abstract scene graphs to represent user intention in fine-grained level. Concrete control signals used in these works can clearly guide model which part of an image to describe. However, only when users are able to see the image can they give such kinds of signals, which is unrealistic for the blind. In this paper, we first propose to use human language as the control signal, which is abstract but more suitable for visually-impaired people to interact with the machine.

\section{Qc-TextCap Dataset}

The Question-controlled Text-aware Image Captioning (Qc-TextCap) task simulates realistic scenario to generate personalized captions for visually impaired people.
An automatic image captioning system firstly generates a general caption $C^{ini}$ about an image $I$ for the user, which describes major objects in the image but contains no scene text information.
Then the user can ask scene text related questions $\mathcal{Q}$ to obtain more specific information.
The system aims to re-generate a new text-aware caption $Y$ to answer questions $\mathcal{Q}$.
However, existing text-aware image captioning datasets (i.e. TextCaps~\cite{DBLP:conf/eccv/SidorovHRS20} and VizWiz-Captions~\cite{DBLP:conf/eccv/GurariZZB20}) only contain <$I, Y$> pairs, which do not support such interactive control of caption generation.
Therefore, in this section, we present an automatic approach to build <$I, C^{ini}, Q, Y$> data samples based on available <$I, Y$> annotations for the Qc-TextCap task.

\subsection{Automatic Datatset Construction}
\label{sec:dataset}
\begin{figure}
    \centering
    \includegraphics[width=0.9\linewidth]{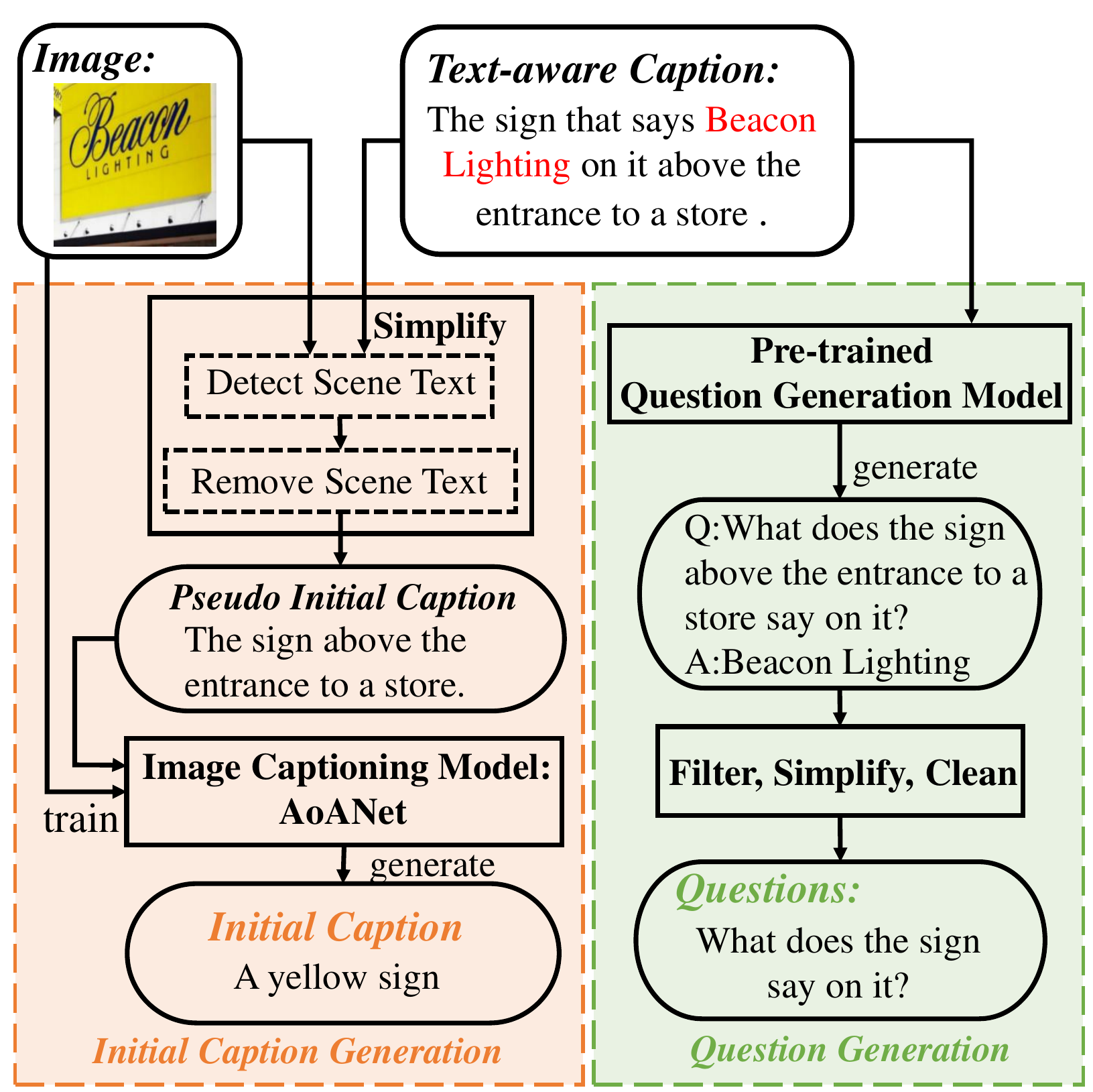}
    \caption{The process of automatic dataset construction for Question-controlled Text-aware Captioning.}
    \label{fig:data_build}
\end{figure}

Figure \ref{fig:data_build} illustrates the pipeline of our automatic dataset construction approach, consisting of \emph{Initial Caption Generation} and \emph{Question Generation} steps.
The \emph{Initial Caption Generation} step generates initial general captions about the image, which only contain visual object information without any scene texts. 
The \emph{Question Generation} step is to produce questions about scene texts in the image.

\iffalse
\begin{figure}
    \centering
    \includegraphics[width=0.9\linewidth]{image/dependency_tree.pdf}
    \caption{Removing scene texts in a text-aware caption with Syntactic Dependency Parsing.}
    \label{fig:dependency}
\end{figure}
\fi

\paragraph{\textbf{1) Initial Caption Generation.}} 
The initial captions should only express essential image contents such as major visual objects without tedious details, especially scene texts in the image.
Though it is possible to generate initial captions via directly applying an image captioning model pre-trained on general captioning datasets such as MSCOCO \cite{DBLP:conf/eccv/LinMBHPRDZ14}, it is not optimal for our task due to domain differences. 
Images in MSCOCO datasets are web images carefully taken by photographers, while we focus on images containing scene texts which may even be taken by the blind. Besides, the initial captions would be better to mention scene text relevant objects for the task.
Hence, we first extract in-domain pseudo initial captions $\tilde{C}^{ini}$ based on $I$ and $Y$,
and use them to train an image captioning model to generate initial caption $C^{ini}$ automatically.

Specifically, we apply two steps to obtain $\tilde{C}^{ini}$: detecting scene texts in $Y$ and then removing the detected scene texts.
Firstly, we detect which words in the text-aware caption $Y$ are scene text words. We perform Optical Character Recognition (OCR) on images via Microsoft Azure Read API\footnote{\url{https://docs.microsoft.com/en-us/azure/cognitive-services/computer-vision/overview-ocr}} and compare the OCR results with words in the caption.
As automatic OCR results are not perfect especially for low-quality images taken by the blind, we further perform a series of Natural Language Processing (NLP) procedures to improve the scene text detection in caption.
If a phrase is recognized as a Named Entity (e.g. a person name or a company name), it is considered as a scene text even though it does not match any OCR results. 
After detecting scene texts in $Y$, we need to remove them without hurting the fluency of the sentence. Naively deleting the words is prone to make grammatical errors. 
Therefore, we use the syntactic dependency parser Spacy\footnote{\url{https://spacy.io/}} to construct a dependency tree and prune the branches containing the scene texts. We present an example to illustrate this process in the supplementary material. In this way, we can obtain pseudo initial captions $\tilde{C}^{ini}$. 

Then we train an automatic image captioning model given $I$ and  $\tilde{C}^{ini}$ pairs on the training set.
Specifically, we use the state-of-the-art AoANet \cite{DBLP:conf/iccv/HuangWCW19} model.
The in-domain training enables AoANet model to generate $C^{ini}$ that mentions major objects in the image. 
Besides, the similarity between $C^{ini}$ and $Y$ is much lower than $\tilde{C}^{ini}$ which is better to simulate real applications where the initial caption may not be perfect. More details about the training of AoANet are provided in our supplementary material.

\paragraph{\textbf{2) Question Generation.}}
The goal is to generate scene text related questions as control signals.
We first use T5 model \cite{DBLP:journals/jmlr/RaffelSRLNMZLL20} trained on SQuADv1 \cite{DBLP:conf/emnlp/RajpurkarZLL16} dataset to generate multiple questions and answers given the target caption $Y$. 
As some QA pairs are not relevant to scene texts, we filter out such QA pairs by checking whether the answers can be found in OCR results. 
Besides, questions about one scene text may contain scene text answers of other questions or extra object descriptions not in initial captions, which leak groundtruth information to the model and should be avoided.
Therefore, we further apply several question cleaning steps to remove such extra leaking information.
As shown in Figure \ref{fig:data_build}, for the initial caption `a yellow sign', a question `what does the sign above the entrance to a store say on it' is transferred into `what does the sign say on it'.
More details about question filtering and question cleaning can be found in the supplementary material.

\subsection{Dataset Analysis}
\begin{table*}
\caption{Datasets statistics of our ControlTextCaps and ControlVizWiz. $Tuple=(I, Y, \tilde{C}^{ini}, C^{ini}, \mathcal{Q})$, where $I$, $Y$, $\tilde{C}^{ini}$, $C^{ini}$ represent image, target caption, pseudo initial caption and automatic initial caption respectively, $\mathcal{Q}=\{Q\}$ represents questions related to the target caption $Y$, $O$ refers to OCR tokens. $N(x)$ is the number of x. $L(x)$ is the average sequence length of x. $P^{obj}(C^{ini})$ means the precision of objects in automatic initial captions.}
\vspace{-10pt}
    \label{tab:dataset_statistic}
    \footnotesize
    %\small
    \centering
    \begin{tabular}{p{0.09\linewidth}p{0.06\linewidth}p{0.05\linewidth}p{0.04\linewidth}p{0.04\linewidth}p{0.04\linewidth}p{0.04\linewidth}p{0.04\linewidth}p{0.05\linewidth}p{0.05\linewidth}|p{0.07\linewidth}p{0.07\linewidth}p{0.06\linewidth}}
    %\begin{tabular}{c|cccccccc}
    \toprule
    Dataset & Split & $N(Tuple)$ & $N(Q)$ & $N(I)$  & $L(Y)$ & $L(Q)$ & $L(O)$ &$L(\tilde{C}^{ini})$ & $L(C^{ini})$ & CIDEr($\tilde{C}^{ini}$) & CIDEr($C^{ini}$) & $P^{obj}(C^{ini})$  \\
    \midrule
    \multirow{3}*{ControlTextcaps} & train & 65,192 & 73,719 & 20,217 & 12.3  & 7.7 & 13.5 & 7.2 & 5.6 & 385.73 & 52.99 & 94.90 \\
     ~ & validation & 4,992 & 5,468 &1,571 & 12.0  & 7.9 & 13.5 & 7.4 & 5.9 & 437.92 & 38.92 & 79.90 \\
     ~ & test & 5,000 & 5,753 &1,429 & 11.9 & 7.5 & 15.1 & 6.8 & 5.3  & 386.11 & 32.73 & 82.04 \\
    \midrule
    \multirow{3}*{ControlVizWiz} & train & 25,339 & 29,139 & 10,763 & 11.8 & 7.4 & 11.3 & 7.0 & 5.2  & 379.32 & 44.34 & 85.25 \\
    ~ & validation & 1,962 & 2,252 & 805 & 11.7 & 7.4 & 11.6 & 6.8 & 5.2 & 387.14 & 32.54 & 70.31 \\
    ~ & test & 1,956 &2,258 & 839 & 11.8 & 7.5 & 12.2 & 7.1 & 5.4 & 406.28 & 32.68  & 66.51\\
    \bottomrule
    \end{tabular}
\end{table*}

\begin{figure}
    \centering
    \includegraphics[width=0.9\linewidth]{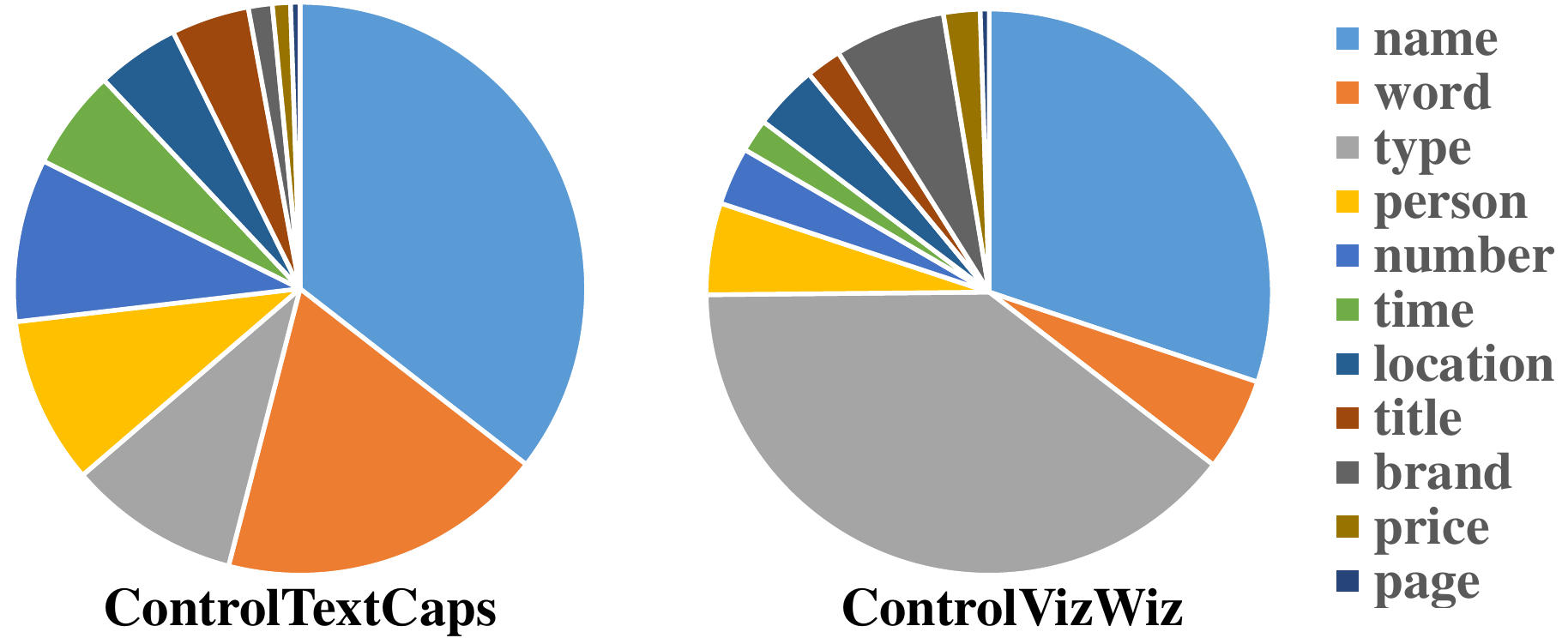}
    \caption{Question type distribution in the ControlTextCaps and ControlVizWiz datasets.}
    \label{fig:question_type}
\end{figure}

\begin{figure}
    \centering
    \includegraphics[width=0.9\linewidth]{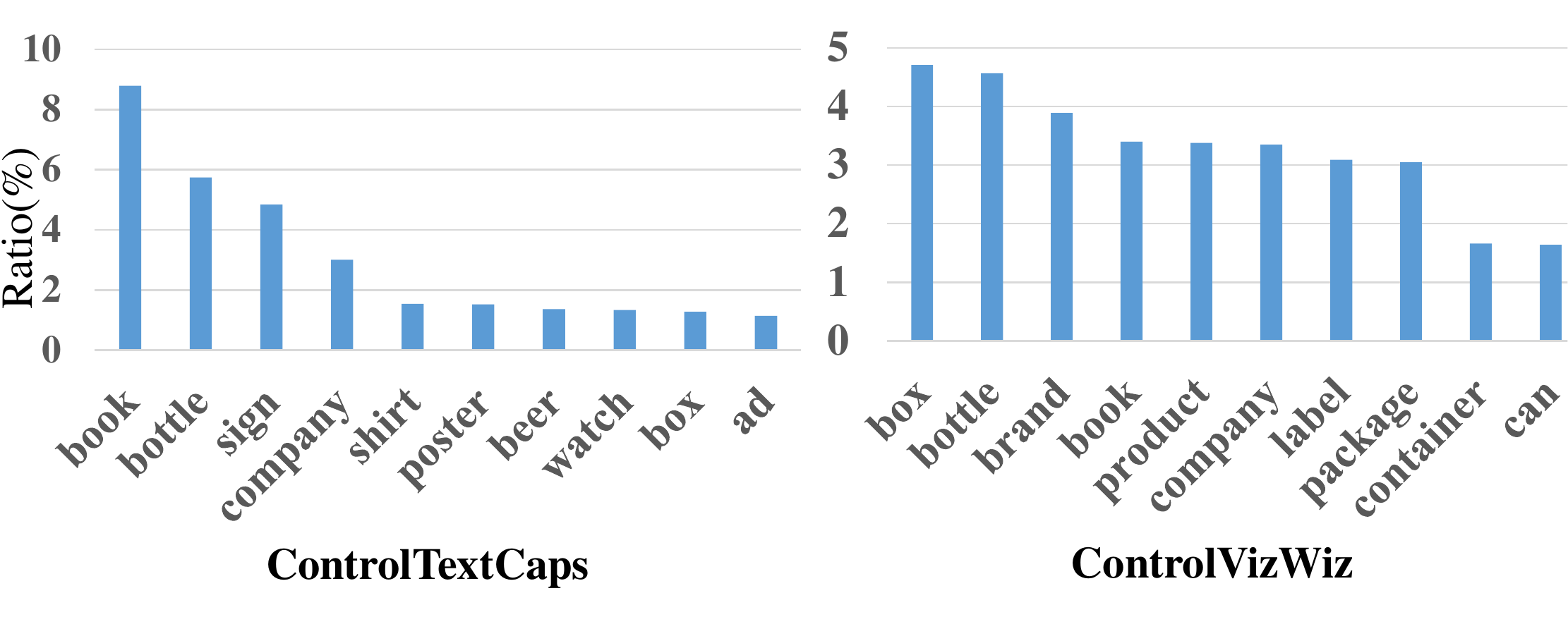}
    \caption{Top 10 objects in the questions of `name' type.}
    \label{fig:top_objects}
\end{figure}

We automatically construct two Qc-TextCap datasets based on TextCaps \cite{DBLP:conf/eccv/SidorovHRS20} and VizWiz-Caption \cite{DBLP:conf/eccv/GurariZZB20}, namely ControlTextCaps and ControlVizWiz. Each example is a tuple of an image ($I$), a target text-aware caption ($Y$), an automatic initial caption ($C^{ini}$), a pseudo initial caption ($\hat{C}^{ini}$), and several questions ($\mathcal{Q}$) towards scene texts in $Y$. The statistics of these two datasets are presented in Table \ref{tab:dataset_statistic}. Each image in raw datasets is annotated with multiple text-aware captions, so an image may appear in multiple tuples in our Qc-TextCap datasets.

\paragraph{\textbf{Initial Caption Quality.}} In terms of sequence length, the text-aware caption $Y$ is around 71\% longer than $\tilde{C}^{ini}$ and 120\% longer than $C^{ini}$. To better quantify the quality of the pseudo initial caption $\tilde{C}^{ini}$ and automatic initial caption $C^{ini}$, we calculate CIDEr \cite{DBLP:conf/cvpr/VedantamZP15} scores of them against the ground truth text-aware captions $Y$ . As shown in Table \ref{tab:dataset_statistic}, $\tilde{C}^{ini}$ achieves very high CIDEr scores, which indicates that pseudo initial captions are very similar with target captions.
Using them as initial captions might not be good for developing the model ability in enriching object description.
The automatic initial caption has much lower CIDEr scores compared to the pseudo initial caption, but achieves acceptable performance in object precision. 
Therefore, the automatic initial caption is more suitable to be used as the initial caption that describes major objects but gives little hint about target text-aware caption. 

\paragraph{\textbf{Question Quality.}}We ask humans to evaluate the quality of automatically generated questions. For each question, we ask a human evaluator to classify it as `No error', `Grammar error' or `Semantic error'. `Grammar error' means the question is indeed asked for scene texts in the corresponding caption but contains some grammar errors. `Semantic error' means the question is irrelevant with the scene text. We ask 10 people to evaluate 2070 questions in ControlTextCaps and 4 people to evaluate 850 questions in ControlVizWiz. According to the statistics, the `Semantic error' only accounts for 13.48\% and 20.35\% in ControlTextCaps and ControlVizWiz, respectively. `No error' accounts for 62.90\% and 55.53\%. This indicates that these automatically generated questions are good enough to support Qc-TextCap task. 

\paragraph{\textbf{Question Diversity.}} Question diversity is critical to simulate different kinds of information needs in real use case. To measure the question diversity of these two datasets, we extract the backbone of questions by Syntactic Dependency Parsing and then classify them to 11 question types by rules. Figure \ref{fig:question_type} presents the question type distribution on two Qc-TextCap datasets. Specifically, questions of `name' type account more than 30\% in both two datasets. To analyze this type of question in a fine-grained way, we further distinguish them according to the queried objects. Figure \ref{fig:top_objects} shows ratio of top 10 queried objects in `name' questions. First, in both datasets, questions about top objects account for a small proportion, which indicates that the objects queried in `name' questions are diverse. Second, top objects queried in ControlTextCaps and ControlVizWiz are different. Questions in ControlVizWiz are more about the objects held by hand, which is consistent with the fact that pictures in VizWiz are all taken by blind people.

\section{QC-TEXTCAP Model}
\begin{figure*}
    \centering
    \includegraphics[width=0.9\linewidth]{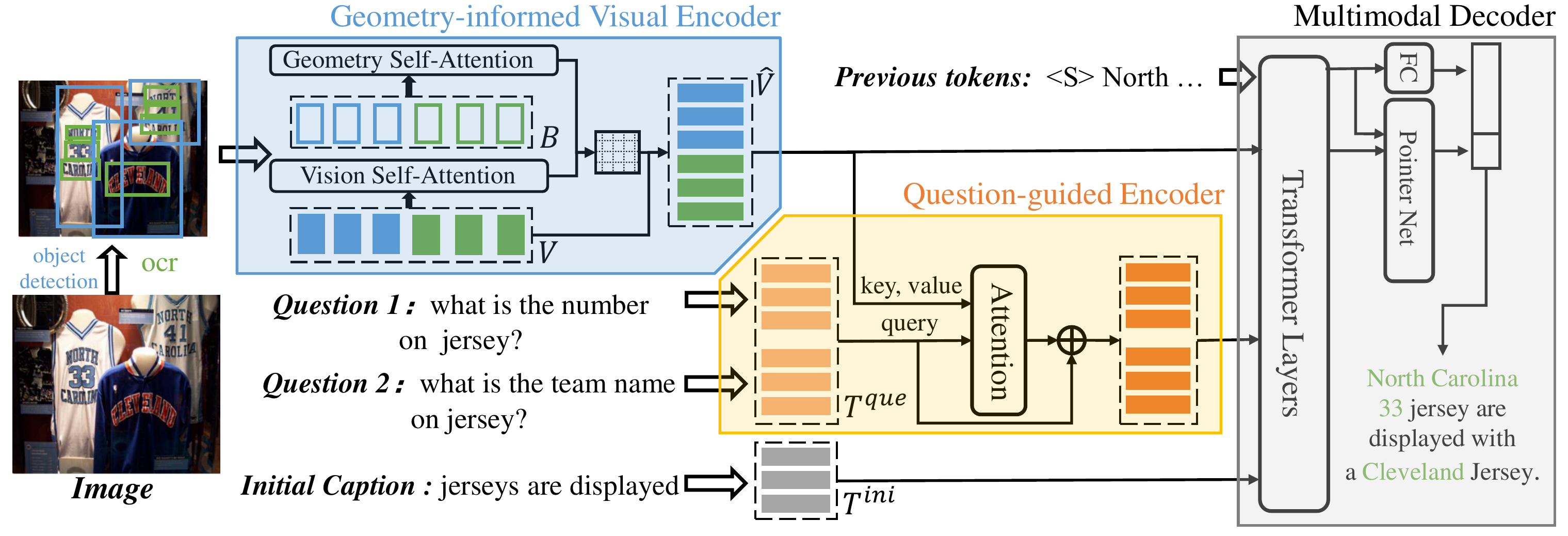}
    \caption{The overall architecture of Geometry and Question Aware Model (GQAM). Geometry-informed Visual Encoder fuses visual object features and scene text features considering their geometry relationships. Question-guided Encoder dynamically selects relevant visual features to questions. Multimodal Decoder takes multimodal features to generate a text-aware caption.}
    \label{fig:overall_model}
\end{figure*}

In this section, we introduce the Geometry and Question Aware Model (GQAM) for the Qc-TextCap task. 
As illustrated in Figure \ref{fig:overall_model}, GQAM consists of three modules, namely Geometry-informed Visual Encoder, Question-guided Encoder and Multimodal Decoder. The Geometry-informed Visual Encoder fuses object region features and scene text region features with relative geometry information.
Question-guided Encoder dynamically selects relevant visual features to encode questions.
The Multimodal Decoder takes inputs of visual, question and initial caption features to sequentially generate a text-aware caption for the question.

Given input $I, C^{ini}, \mathcal{Q}$, we use bottom-up-attention \cite{DBLP:conf/cvpr/00010BT0GZ18} model to detect object bounding boxes $B^{obj}=[b^{obj}_{1}, ..., b^{obj}_{N_{obj}}]$ and Microsoft Azure Read API to detect scene text bounding boxes $B^{ocr}=[b^{ocr}_{1},..., b^{ocr}_{N_{ocr}}]$. Then we extract both object region features $V^{obj}=[v^{obj}_{1}, ..., v^{obj}_{N_{obj}}]$ and scene text region features $V^{ocr}=[v^{ocr}_{1}, ...,v^{ocr}_{N_{ocr}}]$ by bottom-up-attention model. For the initial caption $C^{ini}$ and questions $\mathcal{Q}$, we extract token-level embeddings $T^{ini}=[t^{ini}_{1}, t^{ini}_{2},..., t^{ini}_{N_{ini}}]$ and $T^{que}=[t^{que}_{1},t^{que}_{2}, ..., t^{que}_{N_{que}}]$ (multiple questions in $\mathcal{Q}$ are concatenated to one token sequence) with a trainable three-layer transformer, which is initialized from the first 3 layers of ${\rm Bert_{base}}$ \cite{DBLP:conf/naacl/DevlinCLT19}.

\subsection{Geometry-informed Visual Encoder}
 Spatial relation between object regions and scene text regions is critical for accurately describing the scene text information about an object. For example, for a question `what is the title of the book', the region containing `title' information is certainly included in the region of the `book' object. 
Besides, encoding the relative position between object regions is also beneficial for describing the object relationships in captions. 
Furthermore, due to typesetting, long phrases in images are usually separated into multiple rows or columns. By utilizing the spatial relation between scene text regions, visual features of one phrase can be encoded together. 
Therefore, we apply a module to jointly encode object region features and scene text region features $V=[V^{obj}, V^{ocr}]$ with the help of geometry information $B=[B^{obj}, B^{ocr}]$, namely Geometry Informed Visual Encoder.

The visual encoder is based on multi-head attention \cite{DBLP:conf/nips/VaswaniSPUJGKP17}.
We propose a geometry self-attention mechanism to influence the visual self-attention distribution among both object regions and scene text regions.  The geometry informed self-attention is calculated as follows. For simplicity, we only show the calculation of one head attention in equations.

\begin{gather}
b_{i} = (c^x_i, c^y_i, w_i, h_i)\\
b_{j} = (c^x_j, c^y_j, w_j, h_j)\\
s^{g}_{ij} = {\rm log}(\frac{\lvert c^x_i-c^x_j \rvert}{w_i}, \frac{\lvert c^y_i-c^y_j \rvert}{h_i}, \frac{w_j}{w_i}, \frac{h_j}{h_i})W^{g}, \\
s^{v}_{ij} = v_{i}W^{v}_{Q}(v_{j}W^{v}_{K})^{T},\\
s_{ij} = \frac{s^{g}_{ij}\exp{s^{v}_{ij}}}{\sum^{N}_{k=1}{s^{g}_{ik}\exp{s^{v}_{ik}}}}, \\
\alpha_{i} = [s_{i1}, s_{i2}, ...,s_{iN}], \\
\hat{v_{i}} = \alpha_{i}V,
\end{gather}
where $(c^x,c^y,w,h)$ means (center coordinates, width, height) of a bounding box. $W^{g}, W^{v}_{Q}, W^{v}_{K}$ are learned projection matrices. $b_{i}$ and $v_{i}$ are the bounding box and region feature for the $i^{th}$ region in joint visual region sequence. $s^{g}_{ij}$ and $s^{v}_{ij}$ means the geometry attention score and visual attention score, respectively. $\alpha_{i}$ means the geometry informed attention weight for the $i^{th}$ vision region. $N=N_{obj}+N_{ocr}$. $\hat{v_{i}}$ is the geometry informed visual feature.

\begin{table*}
\caption{Comparison of different models on the ControlTextCaps and ControlVizwiz datasets. `Question' denotes whether the model takes questions as input.}
\vspace{-10pt}
    \label{tab:aoa_simple_cap_experiments}
    \footnotesize
    %\small
    \centering
    \begin{tabular}{p{0.11\linewidth}p{0.11\linewidth}p{0.05\linewidth}|p{0.04\linewidth}p{0.04\linewidth}p{0.04\linewidth}p{0.04\linewidth}p{0.06\linewidth}p{0.06\linewidth}p{0.05\linewidth}p{0.04\linewidth}|p{0.07\linewidth}}
    \toprule
    Dataset & Model  &Question & BLEU1 & BLEU2 & BLEU3 & BLEU4 & METEOR & ROUGE-L & CIDEr & SPICE & AnsRecall \\
    \midrule
    \multirow{4}*{ControlTextcaps} & M4C-Captioner & \XSolidBrush & 34.68 & 21.08 & 13.53 & 8.98 & 15.53 & 32.05 & 102.41 & 20.58 & - \\
    ~ & ControlM4CC & \Checkmark & 52.86 & 40.00 & 30.75 & 23.81 & 25.76 & 48.48 & 215.45 & 37.00 &46.56 \\
    ~ & GQAM w/o GE & \Checkmark & 53.99	& 41.23 & 32.12 & 25.24 & 26.39 & 49.91 & 229.55 & 38.30 &47.14\\
    ~ & GQAM & \Checkmark & \textbf{54.24} & \textbf{41.55} & \textbf{32.50} & \textbf{25.66} & \textbf{26.52} & \textbf{50.07} & \textbf{231.74} & \textbf{38.44} & \textbf{50.92} \\

    \midrule
    \multirow{4}*{ControlVizwiz} & M4C-Captioner & \XSolidBrush & 36.88 & 22.28 & 14.06 & 8.90 & 15.19 & 34.12 & 91.08 & 17.24 & - \\
    ~ & ControlM4CC & \Checkmark & 50.97 & 38.70 & 30.03 & 23.32 & 24.61 & 49.57 & 195.94 & 33.38 & 33.24 \\
    ~  & GQAM w/o GE & \Checkmark & \textbf{53.00} & \textbf{40.67} & \textbf{31.90} & \textbf{25.03} & \textbf{25.25} & \textbf{50.55} & \textbf{210.60} & \textbf{34.58} & 33.39\\
    ~  & GQAM & \Checkmark & 51.61 & 39.62 & 31.06 & 24.33 & 24.82 & 49.73 & 201.35 & 33.81 & \textbf{34.62}\\
    \bottomrule
    \end{tabular}
\end{table*}

\subsection{Question-guided Encoder}
To satisfy multiple information needs in one caption, we propose to progressively find answers for each question from the image and then incorporate answers with the initial caption to produce a final text-aware caption. Thus, it's critical to locate corresponding visual regions for each question before the caption generation.
Besides, for different words in a question, their relevant vision regions should be different. As the example shown in Figure \ref{fig:overall_model}, the word `jersey' is relevant with the `jersey' object regions and the word `number' is relevant with the scene text region `33'.
Taking into account these two points, we design a Question-guided Encoder to dynamically  select relevant visual features for each question at word level before generating text-aware captions. 

With the geometry informed visual features $\hat{V}=[\hat{v}_{1}, ..., \hat{v}_{N}]$ as the key and value, the question token embeddings $T^{que}=[t^{que}_{1},t^{que}_{2}, ..., t^{que}_{N_{que}}]$ as the query, the question-guided attention is calculated as:
\begin{gather}
s^{q}_{ij}=t^{que}_{i}W^{q}_{Q}(\hat{v}_{j}W^{q}_{K})^{T},\\
\beta_{i} = {\rm Softmax}([s^{q}_{i1}, s^{q}_{i2}, ...,s^{q}_{iN}]), \\
t^{v}_{i} = \beta_{i}\hat{V},
\end{gather}
where $W^{q}_{Q}, W^{q}_{K}$ are learned projection matrices, $\hat{v_{j}}$ is the geometry informed visual feature, $\beta_{i}$ means the visual attention weight for the $i^{th}$ token of the question, $t^{v}_{i}$ is the attended visual feature.

To generate a caption that accurately answers a question, key words in questions can always help the decoder describe the relationship between scene texts and objects. For example, for a question `who wrote the book', object word `book' and author name can be fluently connected with a phrase `written by', which can be easily inferred from the keyword `wrote' in the question. So, besides relevant visual features, we retain the token-level text features of questions and combine them with sum operation:
\begin{gather}
\hat{t}^{que}_{i} = t^{v}_{i}W^{q}_{V} + t^{que}_{i}W^{q}_{T},
\end{gather}
where $W^{q}_{V}, W^{q}_{T}$ are learned projection matrices, $\hat{t}^{que}_{i}$ is the multimodal feature for the $i^{th}$ token of the question.

\subsection{Multimodal Decoder}
Multimodal Decoder generates a text-aware caption step by step. At each time step $t$, Multimodal Decoder first fuses multimodal features with muliple transformer layers. The $i^{th}$ transformer layer takes output of the last layer, $[\hat{V}^{obj}_{i-1}, \hat{V}^{ocr}_{i-1}, \hat{T}^{que}_{i-1}, T^{ini}_{i-1}, Y^{dec}_{i-1}]$  , as input, where $\hat{V}^{obj}_{0}$ and $\hat{V}^{ocr}_{0}$ are visual features output by the Geometry-informed Visual Encoder,  $\hat{T}^{que}_{0}$ are token-level question features encoded by the Question-guided Encoder, $T^{ini}_{0}$ are token-level features of the initial caption. $Y^{dec}_{0}$ are embeddings of a fixed-length list storing tokens decoded in previous times steps. With the features output by the final layer of transformers, a pointer network and a fully-connected layer are used to decode the $t^{th}$ token as follows:
\begin{gather}
\hat{y}^{voc}_{t} = {\rm FC}(z^{dec}_{t-1}), \\
\hat{y}^{ocr}_{t,k} = (W^{dec}_{z}z^{dec}_{t-1}+b^{dec})^{T}(W^{ocr}_{z}z^{ocr}_{k}+b^{ocr}), \\
\hat{y}^{ocr}_{t} = [\hat{y}^{ocr}_{t,1}, \hat{y}^{ocr}_{t,2}, ..., \hat{y}^{ocr}_{t,N_{ocr}}],\\
\hat{y}_{t} = {\rm Sigmoid}([\hat{y}^{voc}_{t}, \hat{y}^{ocr}_{t}])
\end{gather}
where $z^{dec}_{t-1}$ is final-layer output of the $(t-1)^{th}$ token in previous decoded token list. $z^{ocr}_{k}$ is the final-layer output of the $k^{th}$ scene text region. $W^{dec}_{z},W^{ocr}_{z}, b^{dec},b^{ocr}$ are trainable parameters. $\hat{y}^{voc}_{t}$ is the predicted score distributed on the fixed vocabulary.  $\hat{y}^{ocr}_{t}$ is the predicted score distributed on the ocr vocabulary made up by scene texts detected from the image. $\hat{y}_{t}$ is the final score distributed on the joint vocabulary.

\subsection{Training}
\noindent\textbf{Training Objective.} During training, we apply the cross-entropy loss for the text-aware caption generation:
\begin{gather}
{\rm Loss} = -\sum_{t=1}^{l}{y_{t} \log(\hat{y}_t)},
\end{gather}
where $y^{t}$ is the ground-truth of the $t^{th}$ token in the target caption $Y$. For a target token appearing in both fixed vocabulary and ocr vocabulary or matching multiple scene text tokens, we randomly sample a possible distribution during training.

\noindent\textbf{Training Strategy.} Given different initial captions as input during training, the model may learn different kinds of abilities. When trained with an initial caption only containing major object words (e.g. automatic initial caption $C^{ini}$) as input, the model should not only improve the image object description (namely description ability) but also add accurate scene text information (namely answering ability) to reduce the generation loss. When trained with a caption similar with ground-truth caption but only lacking scene texts (e.g. pseudo initial caption $\tilde{C}^{ini}$), the model could focus more on the answering ability. Thus, we further explore the influence of different training strategies to these two abilities.
The `pseudo' or `auto' means only using pseudo initial caption $\tilde{C}^{ini}$ or automatic initial caption $C^{ini}$ as initial captions during training, `rand(auto, pseudo)' means randomly choose one of them as the initial caption for each instance during training. Note during inference, we only use automatic initial caption $C^{ini}$ as initial captions because in the real use cases, initial captions are not always in high quality.

\section{Experiments}
In this section, we carry out extensive experiments to evaluate the proposed model on the two constructed Qc-TextCap datasets.

\begin{table*}
\caption{Comparison of different training strategies. `pseudo' or `auto' means only using pesudo initial captions $\tilde{C}^{ini}$ or automatic initial captions $C^{ini}$ as initial captions during training, respectively. `rand(pesudo, auto)' means randomly choosing one of them for each training sample. During inference, only automatic initial captions are used as initial captions.}
\vspace{-10pt}
    \label{tab:training_strategy}
    \footnotesize
    %\small
    \centering
    \begin{tabular}{p{0.10\linewidth}p{0.06\linewidth}p{0.12\linewidth}|p{0.04\linewidth}p{0.04\linewidth}p{0.04\linewidth}p{0.04\linewidth}p{0.04\linewidth}p{0.06\linewidth}p{0.04\linewidth}p{0.04\linewidth}|p{0.07\linewidth}}
    \toprule
    Dataset & Model & train strategy & BLEU1 & BLEU2& BLEU3 & BLEU4 & METEOR & ROUGE-L & CIDEr & SPICE & AnsRecall\\
    \midrule
   \multirow{3}*{ControlTextcaps} & \multirow{3}*{GQAM} & auto & 54.24 & 41.55 & 32.50 & 25.66 & 26.52 & 50.07 & 231.74 & 38.44  & 50.92 \\
    ~ & ~ & pesudo & 43.26 & 29.39 & 20.74 &14.72 & 19.89 & 38.97 & 143.36 & 25.46  & 49.47 \\
    ~ & ~ & rand(auto, pseudo) & \textbf{54.63} & \textbf{42.01} & \textbf{32.96} & \textbf{26.13} & \textbf{26.83} & \textbf{50.50} & \textbf{238.20} & \textbf{38.69}  & \textbf{51.27} \\

    \midrule
    \multirow{3}*{ControlVizwiz} & \multirow{3}*{GQAM} & auto &53.00 & 40.67& 31.90 & 25.03 & 25.25 & 50.55 & 210.60 & 34.58  & 33.39\\
    ~  & ~ & pseudo & 44.85 & 30.56 & 21.70 &15.67& 20.01 & 41.60 & 140.08 & 23.77  & \textbf{34.70} \\
    ~  &  & rand(auto, pseudo) & \textbf{54.41} & \textbf{42.43} & \textbf{33.64} & \textbf{26.79} & \textbf{25.98} & \textbf{51.65} & \textbf{223.23} & \textbf{35.85}  & 33.72 \\
    \bottomrule
    \end{tabular}
\end{table*}

\subsection{Experimental setup}

\noindent\textbf{Baselines.}
1) \textit{Non-controllable baseline: M4C-Captioner} \cite{DBLP:conf/eccv/SidorovHRS20}, which is the state-of-the-art text-aware image captioning model. It fuses object features and scene text features with multi-layer transformers and decodes a caption with a fully connected layer and pointer network. 
2) \textit{Controllable baseline: ControlM4CC}, which extends the M4C-Captioner into a question-controlled captioning model. It has identical architecture with M4C-Captioner, but concatenates initial caption features and question features with object features and scene text features as input.
3) \textit{Controllable variant of GQAM: GQAM w/o GE}, which drops the Geometry-informed Visual Encoder in GQAM to evaluate contributions from different components in the final GQAM model.
 
\noindent\textbf{Evaluation Metrics.}
We use the common captioning evaluation metrics BLEU \cite{DBLP:conf/acl/PapineniRWZ02}, METEOR \cite{DBLP:conf/wmt/DenkowskiL14}, ROUGE-L \cite{lin2004rouge}, CIDEr \cite{DBLP:conf/cvpr/VedantamZP15} and SPICE \cite{DBLP:conf/eccv/AndersonFJG16} to measure the overall quality of generated captions. 
Among these metrics, CIDEr is most suitable for text-aware captioning because it puts more importance on rarer words, especially scene texts. 
To specifically evaluate the question answering ability of question-controlled models, we further calculate the recall of answer tokens as AnsRecall.

\noindent\textbf{Implementation Details.}
On both datasets, we set the max length of the initial caption and the concatenated question as 20. The max length of generated text-aware caption is set to 30. We extract 100 object bounding boxes and at most 50 scene text bounding boxes for each image. 
During training, we set batch size as 50 for both datasets. Training steps are set as 10,000 and 16,000 for ControlVizWiz and ControlTextCaps, respectively.
During inference, we utilize greedy search to decode captions if not specified.

\subsection{Qc-TextCap Performance Evaluation}
We first conduct ablation studies of different models on captioning performance for the Qc-TextCap task.

\textbf{Comparison of non-controllable and controllable models.}
We train all models with the `auto' training strategy to fairly compare different text-aware captioning models. 
As M4C-Captioner is a non-controllable baseline without taking the question as input, we do not calculate the AnsRecall metric of the model. 
As shown in Table \ref{tab:aoa_simple_cap_experiments}, with questions as guidance, question-controlled models achieve much better captioning performance than M4C-Captioner.
Among question-controlled models, our model GQAM outperforms ControlM4CC in both caption generation and question answering, which demonstrates the effectiveness of the overall architecture. 
In both datasets, compared with ControlM4CC, GQAM w/o GE achieves significant improvement, especially on CIDEr scores (+14.1/+13.8). 
This shows that Question-guided Encoder helps locate relevant scene text regions and is critical for describing the relationship between scene texts and objects during generation. 
GQAM achieves better AnsRecall scores than GQAM w/o GE on both datasets, which shows Geometry-informed Visual Encoder indeed helps generate more scene texts by fusing visual region features that are spatially related. 
As for captioning performance, GQAM outperforms GQAM w/o GE on ControlTextCaps but underperforms GQAM w/o GE on ControlVizWiz. This is because that bounding boxes on ControlVizWiz are obviously worse than ControlTextCaps due to the image quality issues. Geometry information of bounding boxes plays a crucial role in Geometry-informed Visual Encoder, so inaccurate bounding boxes can introduce noise into visual features and result in generating some irrelevant scene texts.
For simplicity, we use GQAM to refer to GQAM w/o GE in the following experiments on ControlVizWiz dataset.

\textbf{Comparison of different training strategies.}
Based on the best performed model on each dataset, we further explore the influence of `auto', `pseudo' and `rand(auto, pseudo)' training strategies.
Table \ref{tab:training_strategy} presents the comparison results. 
We find that on test sets of both ControlTextCaps and ContorlVizWiz, models with `pseudo' training strategy achieve similar question answering performance to the ones with `auto' training strategy, but achieve much worse captioning performance. This proves that only using pseudo initial captions is not good for the description ability of models.  Besides, compared with the `auto' training strategy, the `rand(auto, pseudo)' training strategy could further improve both captioning scores and question answering scores on both datasets. This shows that making the model only focus on adding scene text information in part of training steps could strength the answering ability. 

We provide a more detailed ablation study about the contribution of each modality in our supplementary material.

\begin{figure*}
    \centering
    \includegraphics[width=0.92\linewidth]{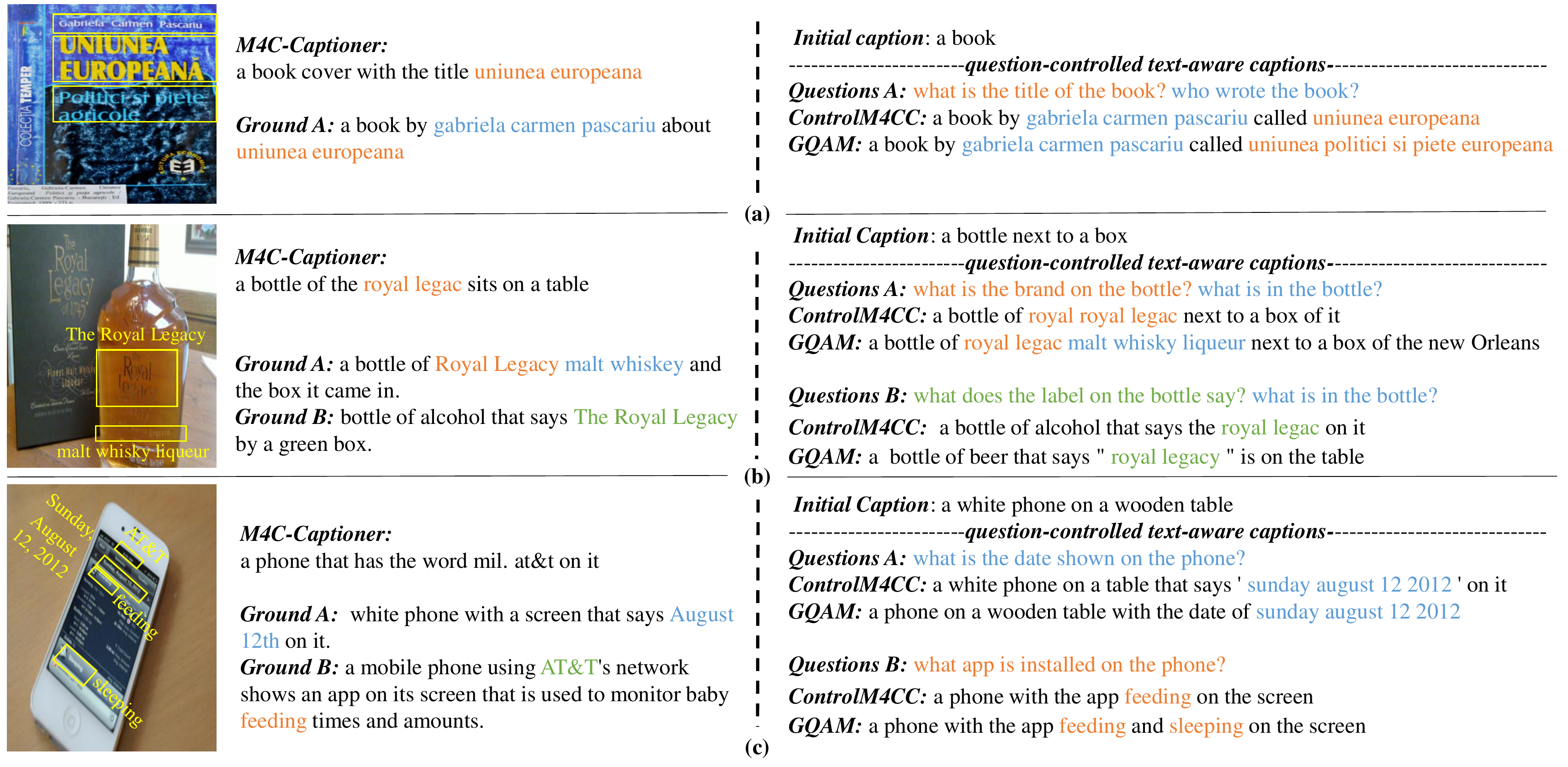}
    \caption{Qualitative results of M4C-Captioner, ControlM4CC and GQAM.}
    \label{fig:Qualitative Results}
\end{figure*}

\begin{table}
    \caption{Diversity evaluation of our GQAM and the text-aware captioning model M4C-Captioner.}
    \label{tab:diversity}
    \vspace{-10pt}
    %\footnotesize
    \small
    \centering
    \begin{tabular}{p{0.23\linewidth}p{0.24\linewidth}p{0.1\linewidth}p{0.1\linewidth}p{0.15\linewidth}}
    \toprule
    Dataset & Model &Div-1 & Div-2 & SelfCIDEr \\
    \midrule
    \multirow{2}*{ControlTextCaps} & M4C-Captioner & 7.44 & 21.11 & 62.58  \\
    ~ & GQAM & \textbf{14.72} & \textbf{38.00} & \textbf{78.32} \\
    \midrule
    \multirow{2}*{ControlVizWiz} & M4C-Captioner & 6.41 & 19.97 & 56.36 \\
    ~ & GQAM & \textbf{10.88} & \textbf{28.71} & \textbf{63.06} \\
    \bottomrule
    \end{tabular}
\end{table}

\begin{table}
    \caption{Human evaluation of accurate scene text information (ST Info) and overall caption quality. For simplicity, we use M4CC to refer to M4C-Captioner}
    \label{tab:human_evaluation}
    \vspace{-10pt}
    %\footnotesize
    \small
    \centering
    \begin{tabular}{cccc}
    \toprule
    Dataset &  & ST Info & Overall Quality \\
    \midrule
    \multirow{3}*{ControlTextcaps} & GQAM$>$M4CC  & 43.48\% & 51.38\% \\
    ~ & GQAM$\approx$M4CC & 42.29\% & 27.67\% \\
    ~ & GQAM$<$M4CC & 14.23\% & 20.95\% \\
    \midrule
    \multirow{3}*{ControlVizWiz} & GQAM$>$M4CC  & 44.30\% & 41.77\% \\
    ~ & GQAM$\approx$M4CC & 39.24\% & 24.05\% \\
    ~ & GQAM$<$M4CC & 16.46\% & 34.18\% \\
    \bottomrule
    \end{tabular}
\end{table}

\subsection{Diversity Evaluation}
Our model is able to generate diverse personalized captions for an image given different questions.
To measure the diversity of these captions, we choose widely used  Div-$n$ \cite{DBLP:conf/cvpr/ChenJWW20,DBLP:conf/iccv/AnejaABS19,DBLP:conf/cvpr/DeshpandeAWSF19} and SelfCIDEr \cite{DBLP:conf/cvpr/WangC19} as diversity metrics. Div-$n$ is the ratio of distinct $n$-gram to the total number of words in the multiple captions for an image. SelfCIDEr is calculated by applying latent semantic analysis on a CIDEr score matrix, which is computed among a set of captions for each image.

For each image, GQAM generates multiple captions with different questions from a relevant question set. 
For the text-aware image captioning model M4C-captioner, we apply beam decoding to get the same number of different captions.
As shown in Table \ref{tab:diversity}, our GQAM significantly outperforms M4C-Captioner on all diversity metrics on both datasets. This indicates that our question control signals indeed guide the model to focus on different image regions and generate personalized captions.

\subsection{Qualitative Evaluation} 
To verify question control signals' contribution in generating more informative captions, we ask 6 human evaluators to evaluate the scene text information and the overall quality of the  text-aware caption. 
For each image, given a caption generated by M4C-Captioner and a caption generated by GQAM, human evaluators are asked to choose 1) which one conveys more scene text information that is accurately described with the object; 2) which one has better overall quality. 
As shown in Table \ref{tab:human_evaluation}, our question-controlled model GQAM accurately describes more scene texts than M4C-Captioner.
Besides, GQAM also performs better in overall quality. This indicates that besides conveying more scene text information, GQAM is also good at organizing them together in natural language.

We show some examples from test sets in Figure \ref{fig:Qualitative Results}. First, general captioning model M4C-Captioner cannot make full use of multiple scene text information in an image. But with multiple questions as control signals, question-controlled text-aware models are guided to focus on multiple scene text regions. For example, as shown in Figure \ref{fig:Qualitative Results}(a), M4C-Captioner only describes the book title without the author name. For question-controlled models, either ControlM4CC or our GQAM successfully outputs these two parts of information given target questions. 
Second, with different questions as control signals, question-controlled models could pay attention to different scene text regions and generate personalized captions. As shown in Figure \ref{fig:Qualitative Results}(c),  when asked about the date on a phone, GQAM  focuses on the `sunday august 12 2012' region, and when asked about the installed application, it describes app names with other scene texts. 
Third, our model GQAM could find relevant scene text regions better than ControlM4CC. In the example presented in Figure \ref{fig:Qualitative Results}(b), given questions about the brand and the content of a bottle, ControlM4CC only describes the brand but GQAM answers both two questions in one caption with the help of the Question-guided Encoder. Further, we find that without background knowledge, GQAM may misunderstand some scene text information. As shown in Figure \ref{fig:Qualitative Results}(c), when asked about the application name on a phone, GQAM focuses on two scene text regions: `feeding' and `sleeping'. These two regions are very similar with applications in shape but are actually two functions of an application according to the text information and background knowledge. More qualitative examples can be found in our supplementary material.

\section{Conclusion}
To generate personalized text-aware captions for visually impaired people, we propose a new challenging task named Question-controlled Text-aware Image Captioning (Qc-TextCap). We use questions about scene texts to control text-aware caption generation due to its convenience in interaction. 
The Qc-TextCap task requires models to comprehend questions, find relevant scene text regions and incorporate answers with an initial caption to produce a final text-aware caption. 
To support this task, we automatically construct datasets ControlTextCaps and ControlVizWiz based on existing text-aware captioning datasets, which will be released publicly. 
We further propose a Geometry and Question Aware Model (GQAM) to progressively encodes relevant visual features and text features. 
On both datasets, GQAM achieves better performance than carefully designed baselines on both captioning and question answering metrics. We further prove that the model with questions as control signals can generate more informative and diverse captions.

\begin{acks}
This work was partially supported by the National Natural Science Foundation of China (No. 61772535 and No. 62072462), Beijing Natural Science Foundation (No. 4192028), National Key R$\&$D Program of China (No. 2020AAA0108600).
\end{acks}

\bibliographystyle{ACM-Reference-Format}
\bibliography{cap}


\begin{thebibliography}{35}

%%% the Web address.

\ifx \showCODEN    \undefined \def \showCODEN     #1{\unskip}     \fi
\ifx \showDOI      \undefined \def \showDOI       #1{#1}\fi
\ifx \showISBNx    \undefined \def \showISBNx     #1{\unskip}     \fi
\ifx \showISBNxiii \undefined \def \showISBNxiii  #1{\unskip}     \fi
\ifx \showISSN     \undefined \def \showISSN      #1{\unskip}     \fi
\ifx \showLCCN     \undefined \def \showLCCN      #1{\unskip}     \fi
\ifx \shownote     \undefined \def \shownote      #1{#1}          \fi
\ifx \showarticletitle \undefined \def \showarticletitle #1{#1}   \fi
\ifx \showURL      \undefined \def \showURL       {\relax}        \fi
% invisible to TeX
\providecommand\bibfield[2]{#2}
\providecommand\bibinfo[2]{#2}
\providecommand\natexlab[1]{#1}
\providecommand\showeprint[2][]{arXiv:#2}

\bibitem[\protect\citeauthoryear{Anderson, Fernando, Johnson, and
  Gould}{Anderson et~al\mbox{.}}{2016}]%
        {DBLP:conf/eccv/AndersonFJG16}
\bibfield{author}{\bibinfo{person}{Peter Anderson}, \bibinfo{person}{Basura
  Fernando}, \bibinfo{person}{Mark Johnson}, {and} \bibinfo{person}{Stephen
  Gould}.} \bibinfo{year}{2016}\natexlab{}.
\newblock \showarticletitle{{SPICE:} Semantic Propositional Image Caption
  Evaluation}. In \bibinfo{booktitle}{\emph{{ECCV} {(5)}}}
  \emph{(\bibinfo{series}{Lecture Notes in Computer Science},
  Vol.~\bibinfo{volume}{9909})}. \bibinfo{publisher}{Springer},
  \bibinfo{pages}{382--398}.
\newblock


\bibitem[\protect\citeauthoryear{Anderson, He, Buehler, Teney, Johnson, Gould,
  and Zhang}{Anderson et~al\mbox{.}}{2018}]%
        {DBLP:conf/cvpr/00010BT0GZ18}
\bibfield{author}{\bibinfo{person}{Peter Anderson}, \bibinfo{person}{Xiaodong
  He}, \bibinfo{person}{Chris Buehler}, \bibinfo{person}{Damien Teney},
  \bibinfo{person}{Mark Johnson}, \bibinfo{person}{Stephen Gould}, {and}
  \bibinfo{person}{Lei Zhang}.} \bibinfo{year}{2018}\natexlab{}.
\newblock \showarticletitle{Bottom-Up and Top-Down Attention for Image
  Captioning and Visual Question Answering}. In
  \bibinfo{booktitle}{\emph{{CVPR}}}. \bibinfo{publisher}{{IEEE} Computer
  Society}, \bibinfo{pages}{6077--6086}.
\newblock


\bibitem[\protect\citeauthoryear{Aneja, Agrawal, Batra, and Schwing}{Aneja
  et~al\mbox{.}}{2019}]%
        {DBLP:conf/iccv/AnejaABS19}
\bibfield{author}{\bibinfo{person}{Jyoti Aneja}, \bibinfo{person}{Harsh
  Agrawal}, \bibinfo{person}{Dhruv Batra}, {and} \bibinfo{person}{Alexander~G.
  Schwing}.} \bibinfo{year}{2019}\natexlab{}.
\newblock \showarticletitle{Sequential Latent Spaces for Modeling the Intention
  During Diverse Image Captioning}. In \bibinfo{booktitle}{\emph{{ICCV}}}.
  \bibinfo{publisher}{{IEEE}}, \bibinfo{pages}{4260--4269}.
\newblock


\bibitem[\protect\citeauthoryear{Biten, Tito, Mafla, i~Bigorda, Rusi{\~{n}}ol,
  Jawahar, Valveny, and Karatzas}{Biten et~al\mbox{.}}{2019}]%
        {DBLP:conf/iccv/BitenTMBRJVK19}
\bibfield{author}{\bibinfo{person}{Ali~Furkan Biten},
  \bibinfo{person}{Rub{\`{e}}n Tito}, \bibinfo{person}{Andr{\'{e}}s Mafla},
  \bibinfo{person}{Llu{\'{\i}}s~G{\'{o}}mez i Bigorda},
  \bibinfo{person}{Mar{\c{c}}al Rusi{\~{n}}ol}, \bibinfo{person}{C.~V.
  Jawahar}, \bibinfo{person}{Ernest Valveny}, {and}
  \bibinfo{person}{Dimosthenis Karatzas}.} \bibinfo{year}{2019}\natexlab{}.
\newblock \showarticletitle{Scene Text Visual Question Answering}. In
  \bibinfo{booktitle}{\emph{{ICCV}}}. \bibinfo{publisher}{{IEEE}},
  \bibinfo{pages}{4290--4300}.
\newblock


\bibitem[\protect\citeauthoryear{Chen, Jin, Wang, and Wu}{Chen
  et~al\mbox{.}}{2020}]%
        {DBLP:conf/cvpr/ChenJWW20}
\bibfield{author}{\bibinfo{person}{Shizhe Chen}, \bibinfo{person}{Qin Jin},
  \bibinfo{person}{Peng Wang}, {and} \bibinfo{person}{Qi Wu}.}
  \bibinfo{year}{2020}\natexlab{}.
\newblock \showarticletitle{Say As You Wish: Fine-Grained Control of Image
  Caption Generation With Abstract Scene Graphs}. In
  \bibinfo{booktitle}{\emph{{CVPR}}}. \bibinfo{publisher}{{IEEE}},
  \bibinfo{pages}{9959--9968}.
\newblock


\bibitem[\protect\citeauthoryear{Cornia, Baraldi, and Cucchiara}{Cornia
  et~al\mbox{.}}{2019}]%
        {DBLP:conf/cvpr/CorniaBC19}
\bibfield{author}{\bibinfo{person}{Marcella Cornia}, \bibinfo{person}{Lorenzo
  Baraldi}, {and} \bibinfo{person}{Rita Cucchiara}.}
  \bibinfo{year}{2019}\natexlab{}.
\newblock \showarticletitle{Show, Control and Tell: {A} Framework for
  Generating Controllable and Grounded Captions}. In
  \bibinfo{booktitle}{\emph{{CVPR}}}. \bibinfo{publisher}{Computer Vision
  Foundation / {IEEE}}, \bibinfo{pages}{8307--8316}.
\newblock


\bibitem[\protect\citeauthoryear{Denkowski and Lavie}{Denkowski and
  Lavie}{2014}]%
        {DBLP:conf/wmt/DenkowskiL14}
\bibfield{author}{\bibinfo{person}{Michael~J. Denkowski} {and}
  \bibinfo{person}{Alon Lavie}.} \bibinfo{year}{2014}\natexlab{}.
\newblock \showarticletitle{Meteor Universal: Language Specific Translation
  Evaluation for Any Target Language}. In \bibinfo{booktitle}{\emph{WMT@ACL}}.
  \bibinfo{publisher}{The Association for Computer Linguistics},
  \bibinfo{pages}{376--380}.
\newblock


\bibitem[\protect\citeauthoryear{Deshpande, Aneja, Wang, Schwing, and
  Forsyth}{Deshpande et~al\mbox{.}}{2019}]%
        {DBLP:conf/cvpr/DeshpandeAWSF19}
\bibfield{author}{\bibinfo{person}{Aditya Deshpande}, \bibinfo{person}{Jyoti
  Aneja}, \bibinfo{person}{Liwei Wang}, \bibinfo{person}{Alexander~G. Schwing},
  {and} \bibinfo{person}{David~A. Forsyth}.} \bibinfo{year}{2019}\natexlab{}.
\newblock \showarticletitle{Fast, Diverse and Accurate Image Captioning Guided
  by Part-Of-Speech}. In \bibinfo{booktitle}{\emph{{CVPR}}}.
  \bibinfo{publisher}{Computer Vision Foundation / {IEEE}},
  \bibinfo{pages}{10695--10704}.
\newblock


\bibitem[\protect\citeauthoryear{Devlin, Chang, Lee, and Toutanova}{Devlin
  et~al\mbox{.}}{2019}]%
        {DBLP:conf/naacl/DevlinCLT19}
\bibfield{author}{\bibinfo{person}{Jacob Devlin}, \bibinfo{person}{Ming{-}Wei
  Chang}, \bibinfo{person}{Kenton Lee}, {and} \bibinfo{person}{Kristina
  Toutanova}.} \bibinfo{year}{2019}\natexlab{}.
\newblock \showarticletitle{{BERT:} Pre-training of Deep Bidirectional
  Transformers for Language Understanding}. In
  \bibinfo{booktitle}{\emph{{NAACL-HLT} {(1)}}}.
  \bibinfo{publisher}{Association for Computational Linguistics},
  \bibinfo{pages}{4171--4186}.
\newblock


\bibitem[\protect\citeauthoryear{Fisch, Lee, Chang, Clark, and Barzilay}{Fisch
  et~al\mbox{.}}{2020}]%
        {DBLP:conf/emnlp/FischLCCB20}
\bibfield{author}{\bibinfo{person}{Adam Fisch}, \bibinfo{person}{Kenton Lee},
  \bibinfo{person}{Ming{-}Wei Chang}, \bibinfo{person}{Jonathan~H. Clark},
  {and} \bibinfo{person}{Regina Barzilay}.} \bibinfo{year}{2020}\natexlab{}.
\newblock \showarticletitle{CapWAP: Image Captioning with a Purpose}. In
  \bibinfo{booktitle}{\emph{{EMNLP} {(1)}}}. \bibinfo{publisher}{Association
  for Computational Linguistics}, \bibinfo{pages}{8755--8768}.
\newblock


\bibitem[\protect\citeauthoryear{Gurari, Zhao, Zhang, and Bhattacharya}{Gurari
  et~al\mbox{.}}{2020}]%
        {DBLP:conf/eccv/GurariZZB20}
\bibfield{author}{\bibinfo{person}{Danna Gurari}, \bibinfo{person}{Yinan Zhao},
  \bibinfo{person}{Meng Zhang}, {and} \bibinfo{person}{Nilavra Bhattacharya}.}
  \bibinfo{year}{2020}\natexlab{}.
\newblock \showarticletitle{Captioning Images Taken by People Who Are Blind}.
  In \bibinfo{booktitle}{\emph{{ECCV} {(17)}}} \emph{(\bibinfo{series}{Lecture
  Notes in Computer Science}, Vol.~\bibinfo{volume}{12362})}.
  \bibinfo{publisher}{Springer}, \bibinfo{pages}{417--434}.
\newblock


\bibitem[\protect\citeauthoryear{Herdade, Kappeler, Boakye, and Soares}{Herdade
  et~al\mbox{.}}{2019}]%
        {DBLP:conf/nips/HerdadeKBS19}
\bibfield{author}{\bibinfo{person}{Simao Herdade}, \bibinfo{person}{Armin
  Kappeler}, \bibinfo{person}{Kofi Boakye}, {and} \bibinfo{person}{Joao
  Soares}.} \bibinfo{year}{2019}\natexlab{}.
\newblock \showarticletitle{Image Captioning: Transforming Objects into Words}.
  In \bibinfo{booktitle}{\emph{NeurIPS}}. \bibinfo{pages}{11135--11145}.
\newblock


\bibitem[\protect\citeauthoryear{Hu, Singh, Darrell, and Rohrbach}{Hu
  et~al\mbox{.}}{2020}]%
        {DBLP:conf/cvpr/HuSDR20}
\bibfield{author}{\bibinfo{person}{Ronghang Hu}, \bibinfo{person}{Amanpreet
  Singh}, \bibinfo{person}{Trevor Darrell}, {and} \bibinfo{person}{Marcus
  Rohrbach}.} \bibinfo{year}{2020}\natexlab{}.
\newblock \showarticletitle{Iterative Answer Prediction With Pointer-Augmented
  Multimodal Transformers for TextVQA}. In \bibinfo{booktitle}{\emph{{CVPR}}}.
  \bibinfo{publisher}{{IEEE}}, \bibinfo{pages}{9989--9999}.
\newblock


\bibitem[\protect\citeauthoryear{Huang, Wang, Chen, and Wei}{Huang
  et~al\mbox{.}}{2019}]%
        {DBLP:conf/iccv/HuangWCW19}
\bibfield{author}{\bibinfo{person}{Lun Huang}, \bibinfo{person}{Wenmin Wang},
  \bibinfo{person}{Jie Chen}, {and} \bibinfo{person}{Xiaoyong Wei}.}
  \bibinfo{year}{2019}\natexlab{}.
\newblock \showarticletitle{Attention on Attention for Image Captioning}. In
  \bibinfo{booktitle}{\emph{{ICCV}}}. \bibinfo{publisher}{{IEEE}},
  \bibinfo{pages}{4633--4642}.
\newblock


\bibitem[\protect\citeauthoryear{Kant, Batra, Anderson, Schwing, Parikh, Lu,
  and Agrawal}{Kant et~al\mbox{.}}{2020}]%
        {DBLP:conf/eccv/KantBASPLA20}
\bibfield{author}{\bibinfo{person}{Yash Kant}, \bibinfo{person}{Dhruv Batra},
  \bibinfo{person}{Peter Anderson}, \bibinfo{person}{Alexander~G. Schwing},
  \bibinfo{person}{Devi Parikh}, \bibinfo{person}{Jiasen Lu}, {and}
  \bibinfo{person}{Harsh Agrawal}.} \bibinfo{year}{2020}\natexlab{}.
\newblock \showarticletitle{Spatially Aware Multimodal Transformers for
  TextVQA}. In \bibinfo{booktitle}{\emph{{ECCV} {(9)}}}
  \emph{(\bibinfo{series}{Lecture Notes in Computer Science},
  Vol.~\bibinfo{volume}{12354})}. \bibinfo{publisher}{Springer},
  \bibinfo{pages}{715--732}.
\newblock


\bibitem[\protect\citeauthoryear{Lin}{Lin}{2004}]%
        {lin2004rouge}
\bibfield{author}{\bibinfo{person}{Chin-Yew Lin}.}
  \bibinfo{year}{2004}\natexlab{}.
\newblock \showarticletitle{Rouge: A package for automatic evaluation of
  summaries}. In \bibinfo{booktitle}{\emph{Text summarization branches out}}.
  \bibinfo{pages}{74--81}.
\newblock


\bibitem[\protect\citeauthoryear{Lin, Maire, Belongie, Hays, Perona, Ramanan,
  Doll{\'{a}}r, and Zitnick}{Lin et~al\mbox{.}}{2014}]%
        {DBLP:conf/eccv/LinMBHPRDZ14}
\bibfield{author}{\bibinfo{person}{Tsung{-}Yi Lin}, \bibinfo{person}{Michael
  Maire}, \bibinfo{person}{Serge~J. Belongie}, \bibinfo{person}{James Hays},
  \bibinfo{person}{Pietro Perona}, \bibinfo{person}{Deva Ramanan},
  \bibinfo{person}{Piotr Doll{\'{a}}r}, {and} \bibinfo{person}{C.~Lawrence
  Zitnick}.} \bibinfo{year}{2014}\natexlab{}.
\newblock \showarticletitle{Microsoft {COCO:} Common Objects in Context}. In
  \bibinfo{booktitle}{\emph{{ECCV} {(5)}}} \emph{(\bibinfo{series}{Lecture
  Notes in Computer Science}, Vol.~\bibinfo{volume}{8693})}.
  \bibinfo{publisher}{Springer}, \bibinfo{pages}{740--755}.
\newblock


\bibitem[\protect\citeauthoryear{Lu, Xiong, Parikh, and Socher}{Lu
  et~al\mbox{.}}{2017}]%
        {DBLP:conf/cvpr/LuXPS17}
\bibfield{author}{\bibinfo{person}{Jiasen Lu}, \bibinfo{person}{Caiming Xiong},
  \bibinfo{person}{Devi Parikh}, {and} \bibinfo{person}{Richard Socher}.}
  \bibinfo{year}{2017}\natexlab{}.
\newblock \showarticletitle{Knowing When to Look: Adaptive Attention via a
  Visual Sentinel for Image Captioning}. In \bibinfo{booktitle}{\emph{{CVPR}}}.
  \bibinfo{publisher}{{IEEE} Computer Society}, \bibinfo{pages}{3242--3250}.
\newblock


\bibitem[\protect\citeauthoryear{Morris, Johnson, Bennett, and Cutrell}{Morris
  et~al\mbox{.}}{2018}]%
        {DBLP:conf/chi/MorrisJBC18}
\bibfield{author}{\bibinfo{person}{Meredith~Ringel Morris},
  \bibinfo{person}{Jazette Johnson}, \bibinfo{person}{Cynthia~L. Bennett},
  {and} \bibinfo{person}{Edward Cutrell}.} \bibinfo{year}{2018}\natexlab{}.
\newblock \showarticletitle{Rich Representations of Visual Content for Screen
  Reader Users}. In \bibinfo{booktitle}{\emph{{CHI}}}.
  \bibinfo{publisher}{{ACM}}, \bibinfo{pages}{59}.
\newblock


\bibitem[\protect\citeauthoryear{Papineni, Roukos, Ward, and Zhu}{Papineni
  et~al\mbox{.}}{2002}]%
        {DBLP:conf/acl/PapineniRWZ02}
\bibfield{author}{\bibinfo{person}{Kishore Papineni}, \bibinfo{person}{Salim
  Roukos}, \bibinfo{person}{Todd Ward}, {and} \bibinfo{person}{Wei{-}Jing
  Zhu}.} \bibinfo{year}{2002}\natexlab{}.
\newblock \showarticletitle{Bleu: a Method for Automatic Evaluation of Machine
  Translation}. In \bibinfo{booktitle}{\emph{{ACL}}}.
  \bibinfo{publisher}{{ACL}}, \bibinfo{pages}{311--318}.
\newblock


\bibitem[\protect\citeauthoryear{Raffel, Shazeer, Roberts, Lee, Narang, Matena,
  Zhou, Li, and Liu}{Raffel et~al\mbox{.}}{2020}]%
        {DBLP:journals/jmlr/RaffelSRLNMZLL20}
\bibfield{author}{\bibinfo{person}{Colin Raffel}, \bibinfo{person}{Noam
  Shazeer}, \bibinfo{person}{Adam Roberts}, \bibinfo{person}{Katherine Lee},
  \bibinfo{person}{Sharan Narang}, \bibinfo{person}{Michael Matena},
  \bibinfo{person}{Yanqi Zhou}, \bibinfo{person}{Wei Li}, {and}
  \bibinfo{person}{Peter~J. Liu}.} \bibinfo{year}{2020}\natexlab{}.
\newblock \showarticletitle{Exploring the Limits of Transfer Learning with a
  Unified Text-to-Text Transformer}.
\newblock \bibinfo{journal}{\emph{J. Mach. Learn. Res.}}  \bibinfo{volume}{21}
  (\bibinfo{year}{2020}), \bibinfo{pages}{140:1--140:67}.
\newblock


\bibitem[\protect\citeauthoryear{Rajpurkar, Zhang, Lopyrev, and
  Liang}{Rajpurkar et~al\mbox{.}}{2016}]%
        {DBLP:conf/emnlp/RajpurkarZLL16}
\bibfield{author}{\bibinfo{person}{Pranav Rajpurkar}, \bibinfo{person}{Jian
  Zhang}, \bibinfo{person}{Konstantin Lopyrev}, {and} \bibinfo{person}{Percy
  Liang}.} \bibinfo{year}{2016}\natexlab{}.
\newblock \showarticletitle{SQuAD: 100, 000+ Questions for Machine
  Comprehension of Text}. In \bibinfo{booktitle}{\emph{{EMNLP}}}.
  \bibinfo{publisher}{The Association for Computational Linguistics},
  \bibinfo{pages}{2383--2392}.
\newblock


\bibitem[\protect\citeauthoryear{Rennie, Marcheret, Mroueh, Ross, and
  Goel}{Rennie et~al\mbox{.}}{2017}]%
        {rennie2017self}
\bibfield{author}{\bibinfo{person}{Steven~J Rennie}, \bibinfo{person}{Etienne
  Marcheret}, \bibinfo{person}{Youssef Mroueh}, \bibinfo{person}{Jerret Ross},
  {and} \bibinfo{person}{Vaibhava Goel}.} \bibinfo{year}{2017}\natexlab{}.
\newblock \showarticletitle{Self-critical sequence training for image
  captioning}. In \bibinfo{booktitle}{\emph{Proceedings of the IEEE Conference
  on Computer Vision and Pattern Recognition}}. \bibinfo{pages}{7008--7024}.
\newblock


\bibitem[\protect\citeauthoryear{Sidorov, Hu, Rohrbach, and Singh}{Sidorov
  et~al\mbox{.}}{2020}]%
        {DBLP:conf/eccv/SidorovHRS20}
\bibfield{author}{\bibinfo{person}{Oleksii Sidorov}, \bibinfo{person}{Ronghang
  Hu}, \bibinfo{person}{Marcus Rohrbach}, {and} \bibinfo{person}{Amanpreet
  Singh}.} \bibinfo{year}{2020}\natexlab{}.
\newblock \showarticletitle{TextCaps: {A} Dataset for Image Captioning with
  Reading Comprehension}. In \bibinfo{booktitle}{\emph{{ECCV} {(2)}}}
  \emph{(\bibinfo{series}{Lecture Notes in Computer Science},
  Vol.~\bibinfo{volume}{12347})}. \bibinfo{publisher}{Springer},
  \bibinfo{pages}{742--758}.
\newblock


\bibitem[\protect\citeauthoryear{Singh, Natarajan, Shah, Jiang, Chen, Batra,
  Parikh, and Rohrbach}{Singh et~al\mbox{.}}{2019}]%
        {DBLP:conf/cvpr/SinghNSJCBPR19}
\bibfield{author}{\bibinfo{person}{Amanpreet Singh}, \bibinfo{person}{Vivek
  Natarajan}, \bibinfo{person}{Meet Shah}, \bibinfo{person}{Yu Jiang},
  \bibinfo{person}{Xinlei Chen}, \bibinfo{person}{Dhruv Batra},
  \bibinfo{person}{Devi Parikh}, {and} \bibinfo{person}{Marcus Rohrbach}.}
  \bibinfo{year}{2019}\natexlab{}.
\newblock \showarticletitle{Towards {VQA} Models That Can Read}. In
  \bibinfo{booktitle}{\emph{{CVPR}}}. \bibinfo{publisher}{Computer Vision
  Foundation / {IEEE}}, \bibinfo{pages}{8317--8326}.
\newblock


\bibitem[\protect\citeauthoryear{Vaswani, Shazeer, Parmar, Uszkoreit, Jones,
  Gomez, Kaiser, and Polosukhin}{Vaswani et~al\mbox{.}}{2017}]%
        {DBLP:conf/nips/VaswaniSPUJGKP17}
\bibfield{author}{\bibinfo{person}{Ashish Vaswani}, \bibinfo{person}{Noam
  Shazeer}, \bibinfo{person}{Niki Parmar}, \bibinfo{person}{Jakob Uszkoreit},
  \bibinfo{person}{Llion Jones}, \bibinfo{person}{Aidan~N. Gomez},
  \bibinfo{person}{Lukasz Kaiser}, {and} \bibinfo{person}{Illia Polosukhin}.}
  \bibinfo{year}{2017}\natexlab{}.
\newblock \showarticletitle{Attention is All you Need}. In
  \bibinfo{booktitle}{\emph{{NIPS}}}. \bibinfo{pages}{5998--6008}.
\newblock


\bibitem[\protect\citeauthoryear{Vedantam, Zitnick, and Parikh}{Vedantam
  et~al\mbox{.}}{2015}]%
        {DBLP:conf/cvpr/VedantamZP15}
\bibfield{author}{\bibinfo{person}{Ramakrishna Vedantam},
  \bibinfo{person}{C.~Lawrence Zitnick}, {and} \bibinfo{person}{Devi Parikh}.}
  \bibinfo{year}{2015}\natexlab{}.
\newblock \showarticletitle{CIDEr: Consensus-based image description
  evaluation}. In \bibinfo{booktitle}{\emph{{CVPR}}}.
  \bibinfo{publisher}{{IEEE} Computer Society}, \bibinfo{pages}{4566--4575}.
\newblock


\bibitem[\protect\citeauthoryear{Vinyals, Toshev, Bengio, and Erhan}{Vinyals
  et~al\mbox{.}}{2015}]%
        {DBLP:conf/cvpr/VinyalsTBE15}
\bibfield{author}{\bibinfo{person}{Oriol Vinyals}, \bibinfo{person}{Alexander
  Toshev}, \bibinfo{person}{Samy Bengio}, {and} \bibinfo{person}{Dumitru
  Erhan}.} \bibinfo{year}{2015}\natexlab{}.
\newblock \showarticletitle{Show and tell: {A} neural image caption generator}.
  In \bibinfo{booktitle}{\emph{{CVPR}}}. \bibinfo{publisher}{{IEEE} Computer
  Society}, \bibinfo{pages}{3156--3164}.
\newblock


\bibitem[\protect\citeauthoryear{Wang, Tang, and Luo}{Wang
  et~al\mbox{.}}{2020}]%
        {DBLP:conf/mm/WangTL20}
\bibfield{author}{\bibinfo{person}{Jing Wang}, \bibinfo{person}{Jinhui Tang},
  {and} \bibinfo{person}{Jiebo Luo}.} \bibinfo{year}{2020}\natexlab{}.
\newblock \showarticletitle{Multimodal Attention with Image Text Spatial
  Relationship for OCR-Based Image Captioning}. In
  \bibinfo{booktitle}{\emph{{ACM} Multimedia}}. \bibinfo{publisher}{{ACM}},
  \bibinfo{pages}{4337--4345}.
\newblock


\bibitem[\protect\citeauthoryear{Wang and Chan}{Wang and Chan}{2019}]%
        {DBLP:conf/cvpr/WangC19}
\bibfield{author}{\bibinfo{person}{Qingzhong Wang} {and}
  \bibinfo{person}{Antoni~B. Chan}.} \bibinfo{year}{2019}\natexlab{}.
\newblock \showarticletitle{Describing Like Humans: On Diversity in Image
  Captioning}. In \bibinfo{booktitle}{\emph{{CVPR}}}.
  \bibinfo{publisher}{Computer Vision Foundation / {IEEE}},
  \bibinfo{pages}{4195--4203}.
\newblock


\bibitem[\protect\citeauthoryear{Wang, Bao, Wu, and Liu}{Wang
  et~al\mbox{.}}{2021}]%
        {DBLP:conf/aaai/WangBWL21}
\bibfield{author}{\bibinfo{person}{Zhaokai Wang}, \bibinfo{person}{Renda Bao},
  \bibinfo{person}{Qi Wu}, {and} \bibinfo{person}{Si Liu}.}
  \bibinfo{year}{2021}\natexlab{}.
\newblock \showarticletitle{Confidence-aware Non-repetitive Multimodal
  Transformers for TextCaps}. In \bibinfo{booktitle}{\emph{{AAAI}}}.
  \bibinfo{publisher}{{AAAI} Press}, \bibinfo{pages}{2835--2843}.
\newblock


\bibitem[\protect\citeauthoryear{Xu, Ba, Kiros, Cho, Courville, Salakhutdinov,
  Zemel, and Bengio}{Xu et~al\mbox{.}}{2015}]%
        {DBLP:conf/icml/XuBKCCSZB15}
\bibfield{author}{\bibinfo{person}{Kelvin Xu}, \bibinfo{person}{Jimmy Ba},
  \bibinfo{person}{Ryan Kiros}, \bibinfo{person}{Kyunghyun Cho},
  \bibinfo{person}{Aaron~C. Courville}, \bibinfo{person}{Ruslan Salakhutdinov},
  \bibinfo{person}{Richard~S. Zemel}, {and} \bibinfo{person}{Yoshua Bengio}.}
  \bibinfo{year}{2015}\natexlab{}.
\newblock \showarticletitle{Show, Attend and Tell: Neural Image Caption
  Generation with Visual Attention}. In \bibinfo{booktitle}{\emph{{ICML}}}
  \emph{(\bibinfo{series}{{JMLR} Workshop and Conference Proceedings},
  Vol.~\bibinfo{volume}{37})}. \bibinfo{publisher}{JMLR.org},
  \bibinfo{pages}{2048--2057}.
\newblock


\bibitem[\protect\citeauthoryear{{Yang}, {Lu}, {Wang}, {Yin}, {Florêncio},
  {Wang}, {Zhang}, {Zhang}, and {Luo}}{{Yang} et~al\mbox{.}}{2020}]%
        {yang2020tap}
\bibfield{author}{\bibinfo{person}{Zhengyuan {Yang}}, \bibinfo{person}{Yijuan
  {Lu}}, \bibinfo{person}{Jianfeng {Wang}}, \bibinfo{person}{Xi {Yin}},
  \bibinfo{person}{Dinei A.~F. {Florêncio}}, \bibinfo{person}{Lijuan {Wang}},
  \bibinfo{person}{Cha {Zhang}}, \bibinfo{person}{Lei {Zhang}}, {and}
  \bibinfo{person}{Jiebo {Luo}}.} \bibinfo{year}{2020}\natexlab{}.
\newblock \showarticletitle{TAP: Text-Aware Pre-training for Text-VQA and
  Text-Caption.}. In \bibinfo{booktitle}{\emph{Proceedings of the IEEE/CVF
  Conference on Computer Vision and Pattern Recognition}}.
  \bibinfo{pages}{8751--8761}.
\newblock


\bibitem[\protect\citeauthoryear{Zheng, Li, and Wang}{Zheng
  et~al\mbox{.}}{2019}]%
        {DBLP:conf/cvpr/ZhengLW19}
\bibfield{author}{\bibinfo{person}{Yue Zheng}, \bibinfo{person}{Yali Li}, {and}
  \bibinfo{person}{Shengjin Wang}.} \bibinfo{year}{2019}\natexlab{}.
\newblock \showarticletitle{Intention Oriented Image Captions With Guiding
  Objects}. In \bibinfo{booktitle}{\emph{{CVPR}}}. \bibinfo{publisher}{Computer
  Vision Foundation / {IEEE}}, \bibinfo{pages}{8395--8404}.
\newblock


\bibitem[\protect\citeauthoryear{Zhu, Gao, Wang, and Wu}{Zhu
  et~al\mbox{.}}{2021}]%
        {DBLP:conf/aaai/ZhuGWW21}
\bibfield{author}{\bibinfo{person}{Qi Zhu}, \bibinfo{person}{Chenyu Gao},
  \bibinfo{person}{Peng Wang}, {and} \bibinfo{person}{Qi Wu}.}
  \bibinfo{year}{2021}\natexlab{}.
\newblock \showarticletitle{Simple is not Easy: {A} Simple Strong Baseline for
  TextVQA and TextCaps}. In \bibinfo{booktitle}{\emph{{AAAI}}}.
  \bibinfo{publisher}{{AAAI} Press}, \bibinfo{pages}{3608--3615}.
\newblock


\end{thebibliography}


\newpage

\setcounter{section}{0}
\renewcommand\thesection{\Alph{section}}
\section{Additional Dataset Construction Details}
\subsection{Scene Text Removing}
\begin{figure}
    \centering
    \includegraphics[width=0.9\linewidth]{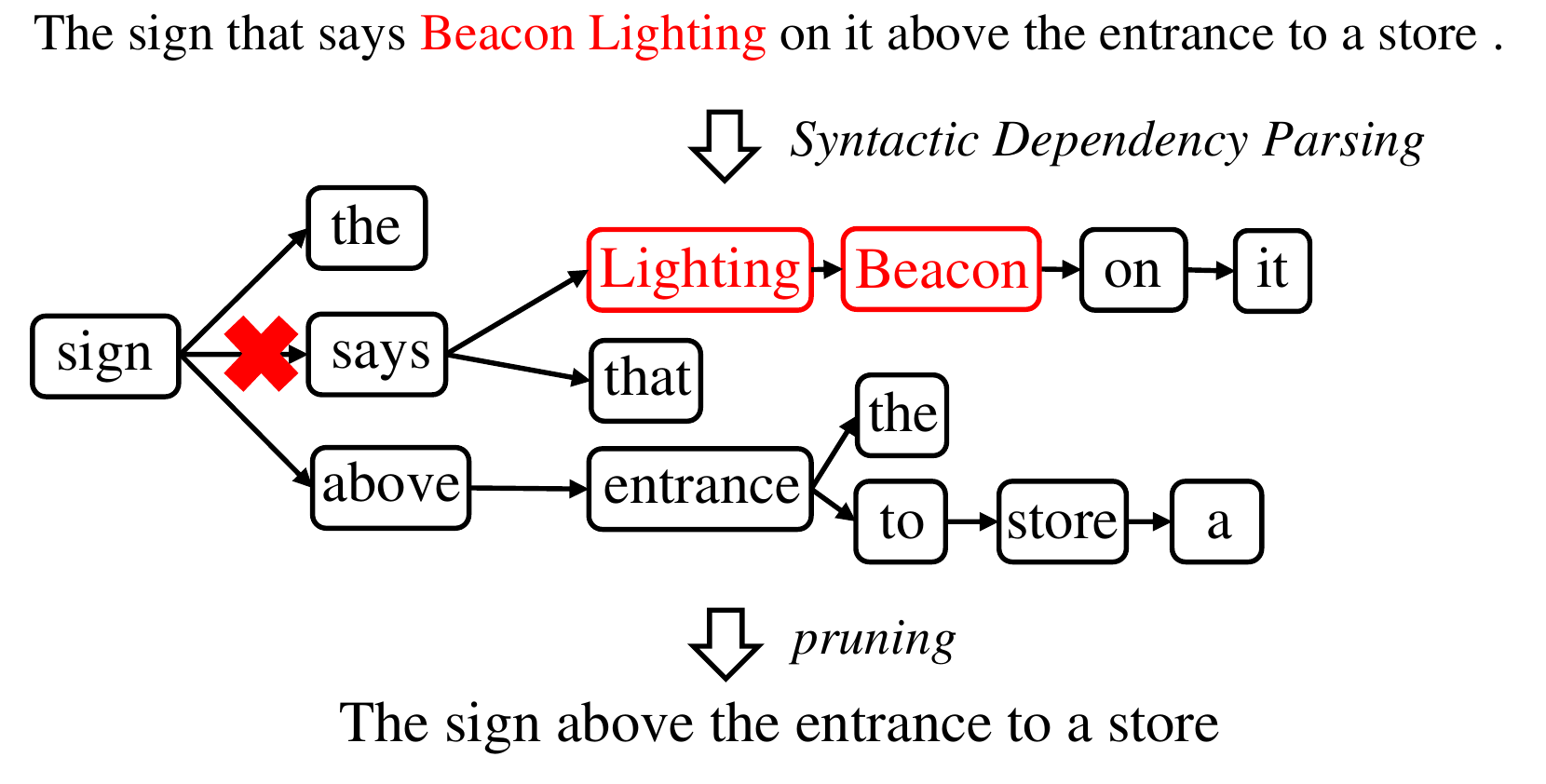}
    \caption{Removing scene texts in a text-aware caption with Syntactic Dependency Parsing.}
    \label{fig:dependency}
\end{figure}

As shown in Figure \ref{fig:dependency}, `Beacon Lighting' are detected scene texts in the text-aware caption. We then perform syntactic dependency parsing to get a dependency tree of the caption. By removing the branch containing scene texts `Beacon' and `Lighting', we get the pseudo initial caption `The sign above the entrance to a store'.

\subsection{Question Filtering}
Questions generated by the pre-trained question generation model are not all about scene texts. For example, for the caption `The sign that says Beacon Lighting on it above the entrance to a store', the model generates a question `where is the sign', which is irrelevant with the scene text information.  We drop this kind of question mainly by judging whether their answers could be found in OCR results. Due to the imperfect of OCR system, a phrase of scene texts may be recognized as multiple OCR tokens. So we keep a question if any token in its answer could be found in OCR results. After the question filtering, only text-ware captions with at least one question towards scene texts are included in our question-controlled datasets.

\subsection{Question Cleaning}
For a text-aware caption containing multiple scene texts, a question asked for one scene text may contain scene text answers of other questions. For example, for a text-aware caption `A book by Mark Hillary titled CEO sits on top of a box', a generated question `what is the title of the book by Mark Hillary' contains the answer of another question `who wrote the book'. Using this kind of questions will leak answers for other questions to the model, so we further remove scene text information in this kind of questions with the simplifying rules introduced in \textbf{\emph{Initial Caption Generation}}.

Besides redundant scene texts, questions generated by the model also contain some extra description information. For a target caption `the sign that says Beacon Lighting on it above the entrance', the question `what does the sign above the entrance say on it' provides extra description information `above the entrance' that is not in the initial caption `yellow sign'. To remove extra description information, we design a template conversion step and backbone extraction step. At template conversion step, We first collect clean questions of high frequency for popular objects, namely template questions, such as `what is the title of the book' and `what is the author of the book' for object `book'. Then we align raw questions to template questions by keywords. For example, if `what book' appears in a raw question, we convert the question to `what is the title of the book'. For questions that are not cleaned by template conversion, we apply a backbone extraction step to remove modifiers by Syntactic Dependency Parsing.

\begin{table}
    \caption{Statistics of training images and captions in Qc-TextCap datasets and text-aware captioning datasets}
    \label{tab:statistic}
    \vspace{-10pt}
    %\footnotesize
    %\small
    \centering
    \begin{tabular}{p{0.3\linewidth}p{0.1\linewidth}p{0.1\linewidth}}
    %\begin{tabular}{ccccccc}
    \toprule
    Dataset & Image & Caption \\
    \midrule
    ControlTextCaps & 20,217 & 65,192 \\
    TextCaps &22,101 & 104,354 \\
    \midrule
    ControlVizWiz & 10,763 & 25,339 \\
    VizWiz-Caption & 28,706 & 121,704 \\
    \bottomrule
    \end{tabular}
\end{table}

\begin{table*}
\caption{Image captioning performance of AoANet models trained on different training data. $P^{obj}(C^{ini})$ means the object precision of generated initial captions. These metrics are all computed at image level. `Split' refers to the train or test split in ControlTextCaps (ControlVizWiz).}
\vspace{-10pt}
    \label{tab:initial_xperiments}
    \footnotesize
    %\small
    \centering
    \begin{tabular}{p{0.10\linewidth}p{0.12\linewidth}p{0.06\linewidth}|p{0.04\linewidth}p{0.04\linewidth}p{0.05\linewidth}p{0.05\linewidth}p{0.04\linewidth}p{0.06\linewidth}p{0.05\linewidth}p{0.04\linewidth}|p{0.06\linewidth}}
    %\begin{tabular}{c|cccccccc}
    \toprule
    Dataset & Train Data & Split  & BLEU1 & BLEU2 & BLEU3 & BLEU4 & METEOR & Rouge-L & CIDEr & SPICE & $P^{obj}(C^{ini})$\\
    \midrule
     \multirow{4}*{ControlTextcaps} & \multirow{2}*{ControlTextcaps} & train & 34.13 & 28.47 & 23.34 & 19.20 & 17.44 & 42.83 & 72.17 & 19.88 & 96.43 \\
    ~ & ~ & test & 24.57 & 17.1 & 11.51 & 7.72 & 11.43 & 31.14	& 26.96 & 11.82 & 81.13 \\
    \cline{2-12}
    ~ & \multirow{2}*{Textcaps} &  Train & 35.96 & 28.36 & 21.72 & 16.47 & 16.11 & 39.99 & 51.84 & 17.56 & 94.40\\
    ~ & ~ & test & 30.3 & 21.41 & 14.56 & 9.70 & 12.41 & 32.65 & 31.18 & 12.63 & 82.04 \\
    \midrule
    \multirow{4}*{ControlVizWiz} &\multirow{2}*{ControlVizWiz} & train & 29.59 & 25.30 & 21.67 & 18.64 & 18.05 & 45.09	& 108.43 & 24.20 & 92.30  \\

    ~ & ~ & test & 16.57 & 10.67 & 6.84 & 4.54 & 8.64 & 27.14 &	23.58 & 9.19 & 53.64 \\
     \cline{2-12}
    ~ & \multirow{2}*{VizWiz-Caption} &  train & 28.00 &	20.11 & 14.10 & 9.78 & 13.07 & 35.46 & 46.05 & 15.53 & 82.25 \\

    ~ & ~ & test & 25.70 & 16.90 & 10.93 & 7.05 & 10.66 & 31.00 & 33.26 & 11.16 & 66.51 \\
    \bottomrule
    \end{tabular}
\end{table*}

\begin{table*}
\caption{Addition Ablation Study. `que', 'ini', `ocr' and 'obj'  refer to the question features $T^{que}$, initial caption features $T^{ini}$, scene text region features $V^{ocr}$ and object region features $V^{obj}$ respectively. Note `w/o ocr' means scene text region features $V^{ocr}$ is not used to fuse with other features but is still used in the pointer network.}
\vspace{-10pt}
    \label{tab:ablation_study}
    \footnotesize
    %\small
    \centering
    \begin{tabular}{p{0.10\linewidth}p{0.09\linewidth}p{0.11\linewidth}|p{0.04\linewidth}p{0.04\linewidth}p{0.05\linewidth}p{0.05\linewidth}p{0.04\linewidth}p{0.06\linewidth}p{0.05\linewidth}p{0.04\linewidth}}
    %\begin{tabular}{c|cccccccc}
    \toprule
    Dataset & Model & Training Strategy & BLEU1 & BLEU2 & BLEU3 & BLEU4 & METEOR & Rouge-L & CIDEr & SPICE\\
    \midrule
     \multirow{6}*{ControlTextcaps} & GQAM w/o que & auto & 37.40 & 23.10 & 14.99 & 9.96 & 16.91 & 33.02 & 109.69 & 21.03 \\
     ~ & GQAM w/o ocr & auto &  53.25 & 40.53 & 31.55 & 24.79 & 25.72 & 49.10 & 221.20 & 37.34 \\ 
     ~ & GQAM w/o obj & auto &  53.95 & 41.30 & 32.23 & 25.38 & 26.38 & 49.72 & 230.11 & 38.13 \\
     ~ & GQAM w/o ini & - & 54.60 & 41.80 & 32.55 & 25.59 & 26.74 & 50.42 &	232.66 & 38.78 \\
     ~ & GQAM & auto & 54.24 & 41.55 & 32.50 & 25.66 & 26.52 & 50.07 & 231.74 & 38.44 \\
     ~ & GQAM & rand(auto, pesudo) &  \textbf{54.63} & \textbf{42.01} & \textbf{32.96} & \textbf{26.13} & \textbf{26.83} & \textbf{50.50} & \textbf{238.20} & \textbf{38.69} \\
    \midrule\ 
     \multirow{6}*{ControlVizWiz} & GQAM w/o que & auto & 36.74 & 22.51 & 14.25 & 9.00 & 15.39 & 34.38 & 93.25 & 18.42  \\
     ~ & GQAM w/o ocr & auto &  50.57 & 38.48 & 29.76 & 23.00 & 23.22 & 48.82 & 187.54 & 31.99 \\ 
     ~ & GQAM w/o obj & auto & 51.99 & 39.65 & 30.95 & 24.17 & 24.69 & 49.66 & 203.67 & 34.32\\
     ~ & GQAM w/o ini & - & 51.84 & 39.44 & 30.71 & 23.99 & 25.26 & 50.20 & 203.29 & 33.99 \\
     ~ & GQAM & auto & 53.00 & 40.67& 31.90 & 25.03 & 25.25 & 50.55 & 210.60 & 34.58 \\
     ~ & GQAM & rand(auto, pesudo) & \textbf{54.41} & \textbf{42.43} & \textbf{33.64} & \textbf{26.79} & \textbf{25.98} & \textbf{51.65} & \textbf{223.23} & \textbf{35.85} \\

    \bottomrule
    \end{tabular}
\end{table*}

\subsection{Training of AoANet}
To generate appropriate initial captions, we train an in-domain general captioning model AoANet given image $I$ and pseudo initial caption $\tilde{C}^{ini}$ pairs. For all captions in raw text-aware captioning datasets (TextCaps/VizWiz-Caption), we produce pseudo initial captions but only part of them are included in our Qc-TextCap datasets due to the question filter, as shown in Table \ref{tab:statistic}. Only using training data of Qc-TextCap datasest (ControlTextCaps/ControlVizWiz) to train an AoANet model  will cause an obvious gap in captioning performance between training split and test split. To alleviate the gap problem, we train the AoANet with all image and pseudo initial caption pairs in TextCaps/VizWiz-Caption training split as training data. As shown in Table \ref{tab:initial_xperiments}, on both datasets, AoANet trained with extra data achieves worse captioning performance on train splits of Qc-TextCap datasets  but better captioning performance on test splits. Especially, AoANet trained with extra data achieves comparable object precision with the one trained with ControlTextCaps or ControlVizWiz on the training split and better object precision on the test split.

\section{Additional Ablation Study}
We perform a more detailed ablation study to show the importance of each modality for Qc-TextCap task. As shown in Table \ref{tab:ablation_study}, all modalities of input are necessary for GQAM. First and foremost, as the control signal, the question contributes most to the captioning performance. Second, by comparing GQAM w/o ini with GQAM trained by `rand(auto, pseudo)' strategy, we find the initial captions is also important in Qc-TextCap task. Finally, to select scene text more accurately, it's necessary to integrate scene text region features with other features.

\begin{figure*}
    \centering
    \includegraphics[width=0.92\linewidth]{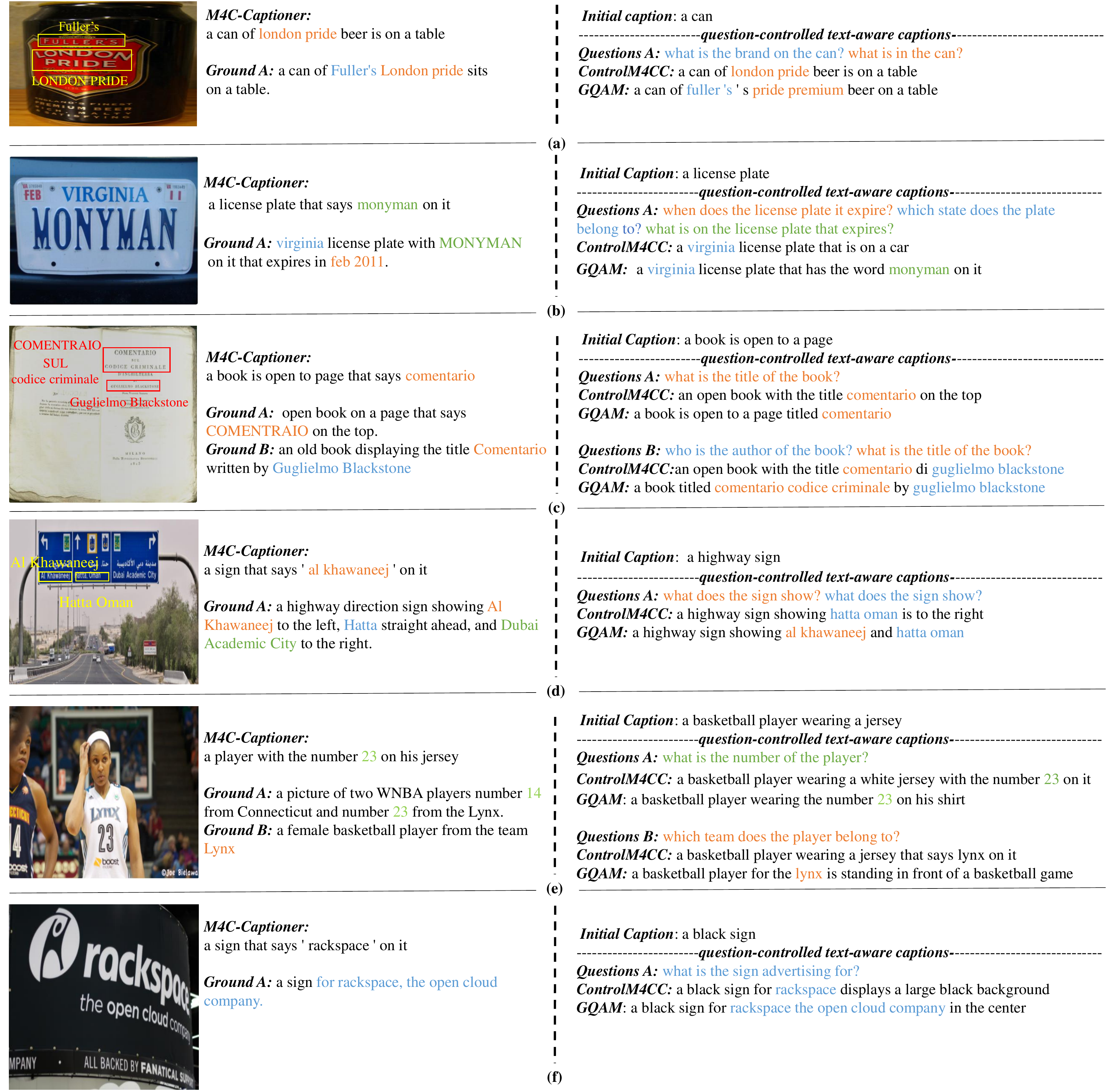}
    \caption{Qualitative Results of M4C-Captioner, ControlM4CC and GQAM.}
    \label{fig:More Qualitative Results}
\end{figure*}

\section{Additional Qualitative Examples}
We present more qualitative examples in Figure \ref{fig:More Qualitative Results}. As shown in Figure \ref{fig:More Qualitative Results}(a) and Figure \ref{fig:More Qualitative Results}(b), our GQAM express more scene text information than M4C-Captioner with multiple questions as control signals. 
Examples in Figure \ref{fig:More Qualitative Results}(c) and \ref{fig:More Qualitative Results}(e) show that, with different questions, question-controlled models ControlM4CC and GQAM generate personalized text-aware captions. In Figure \ref{fig:More Qualitative Results}(c), compared with ControlM4CC, GQAM concatenates the main title and subtitle scene texts to answer the second question, which indicates Geometry-informed Visual Encoder indeed helps recombine a scene text phrase that is separated due to typesetting. Figure \ref{fig:More Qualitative Results}(d) shows the contribution of an good initial caption. M4C-Captioner describes one of scene texts but neglects the type of the sign (`highway sign'), which is important to understand this image. With a good initial caption as input, both ControlM4CC and GQAM retain this key information in their generated captions. Figure \ref{fig:More Qualitative Results}(f) presents that the question could help model to understand the relationship between objects and scene texts. M4C-Captioner just treats `rackspace' as a word on the sign, but GQAM and ControlM4CC understand that it's a company which the sign is advertising for.

%%
%% End of file `sample-sigconf.tex'.

\end{document}